%% file: main.tex
\PassOptionsToPackage{table}{xcolor}
\documentclass[10pt,twocolumn,letterpaper]{article}

 \usepackage[pagenumbers]{cvpr} 
\usepackage{cuted}
\usepackage{caption}
\input{preamble}

\newif\ifincludemain
\newif\ifincludesm
\includemaintrue   
\includesmtrue      

\usepackage{bibunits}
\defaultbibliographystyle{ieeenat_fullname}
\defaultbibliography{main}

\definecolor{cvprblue}{rgb}{0.21,0.49,0.74}
\usepackage[pagebackref,breaklinks,colorlinks,allcolors=cvprblue]{hyperref}

\def\confName{CVPR}
\def\confYear{2027}

\title{ModularReactor: Text-Free Whole-Body Human Reaction Generation}

\author{Hajra Anwar Beg$^{1, 2}$,  Mohamed Daoudi$^{1,2}$,  Angela Bartolo$^{3}$ \\
$^{1}$Univ. Lille, CNRS, Centrale Lille, Institut Mines-Télécom, UMR 9189 CRIStAL, F-59000 Lille, France\\
$^{2}$IMT Nord Europe, Institut Mines-Télécom, Univ. Lille, Centre for Digital Systems, F-59000 Lille, France\\
$^{3}$SCALab, Univ. Lille, CNRS, France\\
{\tt\small  hajra.anwar-beg@imt-nord-europe.fr,mohamed.daoudi@imt-nord-europe.fr}\\{\tt\small 
angela.bartolo@univ-lille.fr }
}

\begin{document}
\maketitle

\vspace{-0.8cm}
\begin{strip}
    \centering
    \includegraphics[width=0.7\linewidth]{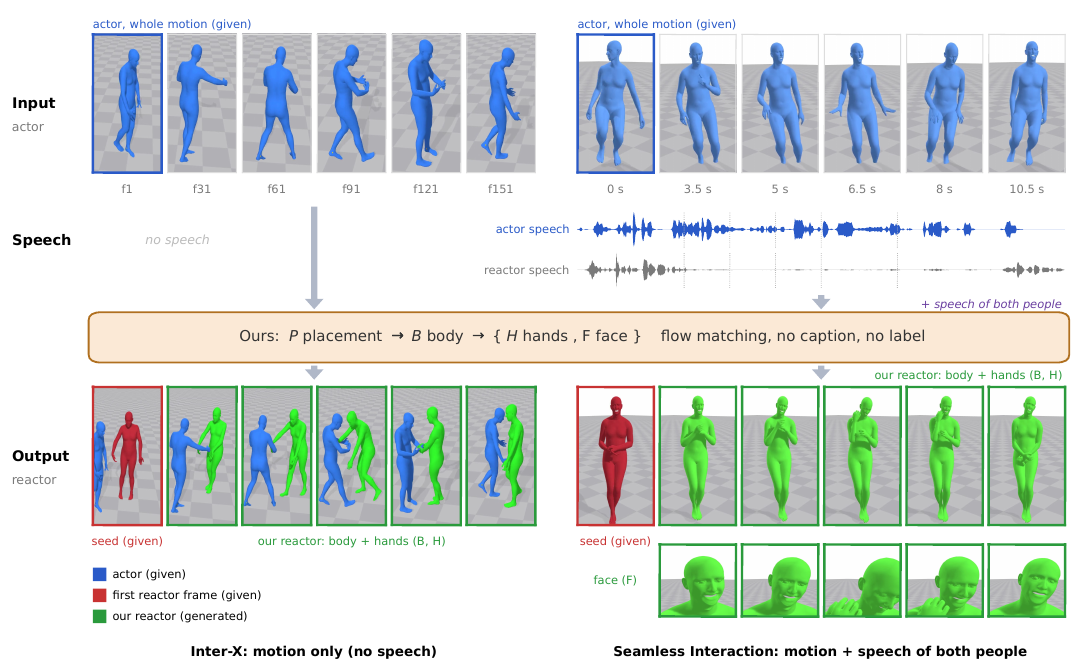}
    \captionsetup{hypcap=false}
    \captionof{figure}{
    \textbf{Text-free physical and conversational reaction generation.}
    Given the complete actor motion  {\color{blue} (blue)} and a reactor initialization
    {\color{red} (red)}, ModularReactor generates the reactor's motion {\color{green} (green)}.
    Left: body and hand reactions on Inter-X.
    Right: body, hand, and facial reactions on Seamless Interaction,
    additionally conditioned on both participants' speech audio.
    }
    \label{fig:thumbnail}
    \captionsetup{hypcap=true}

\end{strip}

\input{sec/0_abstract}    
\input{sec/1_intro}
\input{sec/body}


\input{sec/X_suppl}

{
    \small
    \bibliographystyle{ieeenat_fullname}
    \bibliography{main}
}

\end{document}

%% file: preamble.tex
\usepackage{booktabs}
\usepackage{tabularx}
\usepackage{array}
\usepackage[table]{xcolor}
\usepackage{comment}
\usepackage[T1]{fontenc}

 \newcommand{\mainSecRelated}{2}
\newcommand{\mainSecUserStudy}{4.6}
\newcommand{\mainFigQual}{3}
\newcommand{\mainTabInterXAbl}{3}
\newcommand{\mainTabSlBH}{4}
\newcommand{\mainTabSlFace}{5}

%% file: sec/0_abstract.tex
\vspace{-0.8cm}
\begin{abstract}

Generating a coherent reaction to another person's motion
requires deciding where to move, how to articulate the body
and hands, and, during conversation, how to coordinate facial
behavior with motion and speech.
Existing approaches address these requirements separately:
caption-conditioned methods rely on an explicit interaction
description, while text-free reaction models generate only
a subset of these components.
We propose \emph{ModularReactor}, an offline framework
that addresses both physical and conversational reactions
without captions or action labels at inference. Its conditional hierarchy,
$P \rightarrow B \rightarrow \{H,F\}$,
generates placement, then body motion, followed by hands
and face, according to the available dataset modalities.
Each flow-matching module infers its own intent embedding
from observed signals, using semantic supervision only
during training.
On Inter-X, our model achieves competitive FID on the whole test set
(0.054 versus 0.061) and lower MPJPE on 400 test clips
(27.2 versus 42.2\,cm) than the strongest
caption-conditioned baseline.
It obtains the lowest mean FID among the compared methods
on InterHuman (1.66) and generates body, hand,
and facial reactions on Seamless Interaction.
Ablations demonstrate the benefit of modular generation,
while caption-conditioned baselines retain stronger
text--motion alignment.
\end{abstract}


%% file: sec/1_intro.tex
\vspace{-0.4cm}
\section{Introduction}
\label{sec:intro}

Generating human reactions to a partner's motion is essential
for character animation, video games, and virtual agents. A convincing reaction coordinates spatial placement, body motion, and hand articulation, together with facial expressions
and jaw motion during conversations.
These components are coupled: placement constrains body motion,
while hands and face must remain consistent with the body
and interaction context. Modeling these dependencies is central to whole-body
reaction generation.

Existing approaches address subsets of this problem.
Text-driven interaction models generate both participants
from a description~\cite{liang2024intergen,ruizponce2024in2in,
javed2025intermask,ruizponce2026interact2ar}.
Several strong reaction-generation methods instead condition
on actor motion and an interaction
caption~\cite{xu2025moreact,geng2026armflow}.
Although captions provide semantic guidance, they require
a description for each sequence, motivating the inference of reaction-relevant information directly from motion
and, when available, speech.

Text-free reaction models~\cite{ghosh2024remos,
cen2025readytoreact,beg2026reactionmamba} generate body
and sometimes hand motion, but not facial behavior.
ReMoS~\cite{ghosh2024remos} and
ReactionMamba~\cite{beg2026reactionmamba} output joint
positions rather than articulated rotations.
Conversational models~\cite{mughal2024convofusion,
ng2024audio2photoreal,peng2026dyadit,nagano2026dyaplex}
generate gestures and sometimes facial motion from
speech-related inputs, but primarily target conversations
rather than nonverbal physical interactions.
In Supplementary material we compare their
inputs, initialization requirements, and outputs.
These complementary settings motivate a common framework
for physical and conversational reactions.

We introduce \emph{ModularReactor}, an offline framework
that generates whole-body reactions from complete actor
motion, reactor initialization, and, for conversations,
both participants' recorded speech audio.
It requires no interaction caption or externally supplied
action or intent label at inference.
Module-specific initialization includes the initial pose
and, where used, velocity information.

This setting targets applications in which the reactor's
initial state is given and its subsequent motion must be
generated in response to the observed actor motion.

Four modules generate spatial placement $P$, body motion $B$,
hand articulation $H$, and facial behavior $F$ through
the hierarchy $P \rightarrow B \rightarrow \{H,F\}$:
placement conditions the body, which conditions hands
and face.
Each module is a conditional flow-matching
network~\cite{lipman2023flow,liu2023rectified}
using bidirectional Mamba
blocks~\cite{gu2023mamba,zhu2024vim}.
Modules share a design but have separate parameters
and are trained independently using ground-truth (GT) upstream motion, which is replaced by generated predictions at inference.

Each module predicts interaction intent from its conditioning inputs
and maps the class probabilities to a learned embedding.
Training supervision uses dataset categories,
caption-derived labels, or conversation prompts;
inference uses the modules' predictions.
Thus, \emph{text-free} describes inference,
while training retains semantic supervision.

\vspace{-0.05cm}
Without captions, ModularReactor achieves a FID
(Fr\'echet inception distance) comparable to the best
caption-conditioned model on Inter-X
(0.054 versus 0.061 for our rerun of ReMFlow~\cite{geng2026armflow}, 0.058 as published),
with lower joint position and velocity errors.
On InterHuman, it obtains the lowest mean FID among
compared methods, while caption-conditioned models
retain higher text--motion retrieval scores (R-precision).
On Seamless Interaction, our model achieves a lower body Fréchet gesture distance (FGD) than that of ConvoFusion~\cite{mughal2024convofusion}
retrained on the same windows (0.38 versus 1.79--2.67).
Facial reaction correlation (FRC) with references
to the same prompt is higher than for our adapted
Trans-VAE baseline from
REACT~\cite{song2023react2023,song2025react}
(0.148 versus 0.010).
Our faces also vary more over time, whereas
Trans-VAE produces nearly static faces but obtains
a lower facial reaction distance (FRD).

\textbf{Contributions.}
Our main contributions are:
\vspace{-0.05cm}
\begin{itemize}

    \item \textbf{A framework for text-free whole-body
    reaction generation.}ModularReactor generates placement, body, hands,
    and face according to dataset modalities,
    supporting physical interactions and conversations
    without captions or externally supplied intent
    labels at inference.

    \item \textbf{A conditional modular hierarchy
    with inferred intent.}
    Independently trained flow-matching modules follow
    $P \rightarrow B \rightarrow \{H,F\}$,
    with module-specific intent predictors.
    Ablations support the decomposition's contribution
    to joint position and velocity accuracy.

    \item \textbf{Evaluation across physical and
    conversational interactions.}
    We evaluate on Inter-X, InterHuman, and a curated
    Seamless Interaction protocol with transcript-verified
    roles and disjoint conversation pairs and recordings
    across splits.
    Comparisons include published results, checkpoint
    reruns, and retrained baselines, with ablations
    and confidence intervals over repeated generation runs.

\end{itemize}

%% file: sec/body.tex
\section{Related Work}
\label{sec:related}

\noindent\textbf{From interaction generation to motion-conditioned reactions.}
Text-conditioned interaction models, including
InterGen~\cite{liang2024intergen},
in2IN~\cite{ruizponce2024in2in},
InterMask~\cite{javed2025intermask}, and
Interact2Ar~\cite{ruizponce2026interact2ar},
generate both participants from a shared description.
Reaction generation instead observes an actor
and synthesizes a responding
participant~\cite{chopin2023interformer,
xu2024regennet,zhang2025reactffusion,jiang2025arflow}.
Within this setting,
MoReact~\cite{xu2025moreact} and ReMFlow~\cite{geng2026armflow} additionally
require an interaction caption.
UMF~\cite{huang2026umf} likewise supports
caption-conditioned sequential generation of
two participants; we adapt its second-person
generation stage to an observed actor.
ModularReactor addresses the complementary question
of generating an articulated reaction from observed
motion and reactor initialization, without an
externally supplied description of the interaction.

\noindent
\textbf{Decomposing articulated reactions.}
Generating body and hand motion without supplied
text has been explored by
ReMoS~\cite{ghosh2024remos} and
ReactionMamba~\cite{beg2026reactionmamba}.
ReMoS already uses a body-to-hands cascade,
while ReactionMamba combines a motion autoencoder
with state-space modeling.
MoReact~\cite{xu2025moreact} separately generates
trajectory and body motion, and
Interact2Ar~\cite{ruizponce2026interact2ar}
uses parallel trajectory, body, and hand branches.
Our contribution builds on these decomposition
strategies through the conditional hierarchy
$P\rightarrow B\rightarrow\{H,F\}$:
placement constrains body generation, and the
generated body conditions both finger articulation
and facial behavior.
The modules are trained independently and composed
at inference, with module-specific inferred intent.
Unlike the joint-position outputs of ReMoS and
ReactionMamba, our body and hand representations
include articulated rotations.
Think-Then-React~\cite{tan2025ttr} also avoids
user-supplied text, but generates intermediate
textual descriptions; our intent representation
is inferred without generating text.

\noindent
\textbf{Physical and conversational reactions.}
Conversational models address complementary
aspects of reactive behavior.
ConvoFusion~\cite{mughal2024convofusion},
DyaDiT~\cite{peng2026dyadit}, and
ReactMotion~\cite{luo2026reactmotion}
generate gestures or listener motion from
speech-related inputs, with partner-motion
conditioning available in DyaDiT.
Audio2Photoreal~\cite{ng2024audio2photoreal}
generates body, hands, and face from conversational
audio, while DyaPlex~\cite{nagano2026dyaplex}
jointly generates speech and motion conditioned
on the partner.
UniLS~\cite{chu2025unils} and the
REACT challenges~\cite{song2023react2023,
song2025react} focus on facial and head behavior
or facial reactions.
ModularReactor connects physical and conversational
reaction generation through a common modular design:
motion conditions physical reactions, while
conversational generation additionally uses both
participants' recorded speech.
Its facial module receives the generated reactor
body as well as partner observations and audio.
The modules are instantiated and trained separately
according to the modalities available in each dataset.

\noindent\textbf{Offline versus online reaction generation.}
Ready-to-React~\cite{cen2025readytoreact},
ARMFlow~\cite{geng2026armflow}, and ReMoGen~\cite{ye2026remogen}
generate reactions from the motion observed so far,
which supports interactive use.
Like ReMFlow~\cite{geng2026armflow},
ReGenNet~\cite{xu2024regennet}, and
InterFormer~\cite{chopin2023interformer},
ModularReactor assumes that the actor's complete motion
sequence is available before synthesis, as in character
animation from recorded motion.
Each module uses bidirectional Mamba blocks so that
the reaction at a given frame can depend on both
preceding and subsequent actor motion.

\section{ModularReactor}
\label{sec:method}

\subsection{Problem formulation and modular generation}
\label{sec:problem}
\label{sec:hierarchy}

We consider offline human reaction generation. Given the complete actor motion $A=\{a_i\}_{i=1}^{T}$, a reactor initialization $S$,
and optional speech features $U=(U^a,U^r)$ from both participants, we aim to sample a plausible reactor sequence coordinated with the actor.
All conditioning sequences are available before generation.
No caption or externally supplied action or intent label is required at inference; semantic annotations supervise intent prediction only during training
(Section~\ref{sec:intent}).
The initialization $S$ provides module-specific conditioning rather than a hard constraint: all reactor frames are generated.

 We represent the reactor sequence as $Y=(O,B,H,F)$, comprising spatial placement (O), body motion (B), hand motion (H), and facial behavior (F).
Placement is expressed through root offsets $O=\{o_i\}_{i=1}^{T}$, where $o_i=r_i^r-r_i^a$
denotes the reactor root position relative to the actor in a common coordinate system.
To capture dependencies between these components, we define the conditional model, with
$C=(A,S,U)$,
as
\begin{equation}
\begin{aligned}
p_\theta(O,B,H,F \mid C)
={}& p_{\theta_P}(O \mid C)\\
&\times p_{\theta_B}(B \mid O,C)\\
&\times p_{\theta_H}(H \mid B,C)\\
&\times p_{\theta_F}(F \mid B,C).
\end{aligned}
\label{eq:factorization}
\end{equation}
This factorization establishes the hierarchy $P \rightarrow B \rightarrow \{H,F\}$:
placement conditions body generation, and the generated body subsequently conditions the hands
and face.
The hand and face modules are conditionally independent given the body and observed conditions,
and receive no separate placement stream. Each module uses the subset of observations specified
below; in particular, $P$ does not use speech.

The modules are trained independently with ground-truth upstream sequences, which are replaced by generated predictions at inference.

\noindent
\textbf{Dataset-specific instantiations.}
The framework is instantiated according to the motion components available in each dataset, with parameters trained separately per dataset.
On Inter-X, we use $P$, $B$, and $H$.
On InterHuman, we use $P$ and $B$.
On Seamless Interaction, placement is fixed and we use $B$, $H$, and $F$, conditioned additionally on the speech
audio of both participants.

\begin{figure}[t]
\centering
\includegraphics[width=0.84\linewidth]{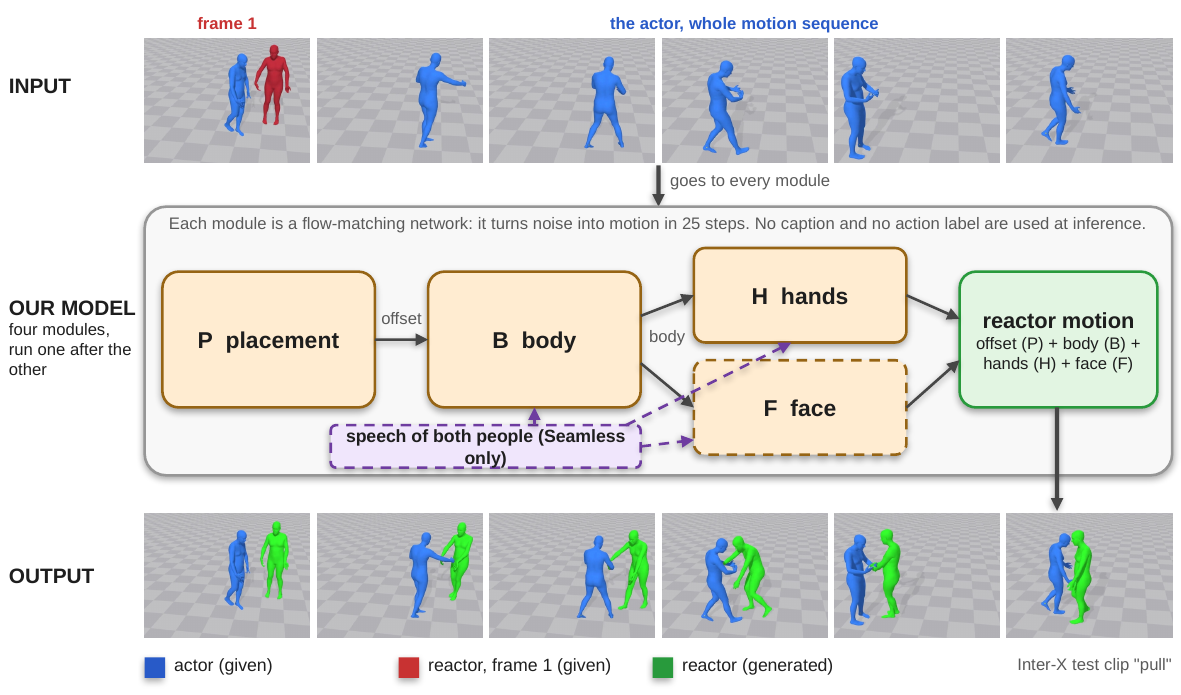}
    \vspace{-0.4cm}
\caption{\textbf{Overview of ModularReactor.}
Given the complete actor motion and a reactor initialization, our framework generates reactions through the hierarchy
$P \rightarrow B \rightarrow \{H,F\}$. The placement module $P$ predicts root offsets that condition
body generation $B$; the generated body then conditions hand articulation $H$ and facial behavior $F$.
Bottom: generated reaction for an Inter-X ``pull'' sequence.
{\color{blue} (blue)}: observed actor; {\color{red} (red)}: reactor initialization;
green: generated reactor.}
    \vspace{-0.5cm}
\label{fig:overview}
\end{figure}

\subsection{Representation}
\label{sec:rep}

We use $i$ for frame indices and $t$ for flow time.
Both participants are expressed in a common coordinate system centered at the actor's initial root position.
Placement is represented by reactor-to-actor root offsets, as defined in Section~\ref{sec:problem}.
Following ARDY~\cite{zhao2026ardy} and HumanML3D~\cite{guo2022humanml3d}, body features combine pelvis-relative joint positions, continuous 6D
rotations~\cite{zhou2019continuity}, joint velocities, and foot-contact indicators:
$\mathbf{b}_i\in\mathbb{R}^{265}$.
The 30 finger joints use 6D rotations,
$\mathbf{h}_i\in\mathbb{R}^{180}$, yielding
$\mathbf{m}_i=[\mathbf{b}_i\,\|\,\mathbf{h}_i]
\in\mathbb{R}^{445}$ in the SMPL-X
layout~\cite{pavlakos2019smplx}, where $[\cdot\,\|\,\cdot]$ denotes concatenation.
Facial features comprise 50 FLAME expression coefficients~\cite{li2017flame} and three jaw-rotation
parameters, $\mathbf{f}_i\in\mathbb{R}^{53}$, with an additional validity flag.
For conversations, each participant's audio is encoded using wav2vec2~\cite{baevski2020wav2vec},
temporally aligned with motion, and reduced to 64 dimensions by PCA fitted only on training data.
Full representation and preprocessing details are provided in supplementary material.

\subsection{Flow matching}
\label{sec:fm}

Each module uses conditional flow
matching~\cite{lipman2023flow,liu2023rectified}
with v-prediction~\cite{salimans2022progressive}.
Given a standardized target $\mathbf{x}_1$,
independent noise
$\mathbf{x}_0\sim\mathcal{N}(\mathbf{0},\mathbf{I})$,
and $t\sim\mathcal{U}(0,1)$, we define
\begin{equation}
\begin{aligned}
    \mathbf{x}_t
    &= (1-t)\mathbf{x}_0+t\mathbf{x}_1,\\
    \mathbf{v}_t
    &= t\mathbf{x}_0-(1-t)\mathbf{x}_1.
\end{aligned}
\end{equation}
The conditional network is trained with
\begin{equation}
    \mathcal{L}_{\mathrm{FM}}
    =
    \mathbb{E}\!\left[
        \|g_\theta(\mathbf{x}_t,t,\mathbf{c})
        -\mathbf{v}_t\|_2^2
    \right],
\end{equation}
averaged over valid frames, where $\mathbf{c}$ denotes the module's conditioning information.
At inference, v-predictions are converted into integration velocities.
Generation uses 25 sampling steps, with Euler updates followed by a final clean-sample estimate.
Conversion formulas and sampling details are provided in the supplementary material.

\subsection{Shared network}
\label{sec:network}

The four modules use a common network architecture (figure in the supplementary material), with separate parameters and module-specific target representations, conditioning inputs, and network dimensions.

Following ARDY~\cite{zhao2026ardy}, we group consecutive frames into non-overlapping pairs to form temporal
tokens for both the generated and actor motion streams. Temporal dependencies are modeled using bidirectional
Mamba blocks:
\begin{equation}
    \operatorname{BiMamba}(\mathbf{h})
    =
    \operatorname{LN}\!\left(
        M_{\rightarrow}(\mathbf{h})
        +
        M_{\leftarrow}(\mathbf{h})
    \right),
\end{equation}
where $\operatorname{LN}$ denotes layer normalization. Each directional branch consists of a
Mamba layer~\cite{gu2023mamba} followed by a feed-forward layer, both using pre-normalization and residual
connections. The forward branch processes tokens in chronological order, while the backward branch processes them in
reverse order and restores their original ordering before fusion. This bidirectional processing gives each token access to both past and future context~\cite{zhu2024vim}, consistent with our offline generation setting.

\noindent\textbf{Intent prediction.}
\label{sec:intent}
A conditioning encoder maps the module’s input streams to tokens K. An intent predictor processes these tokens and is supervised by a cross-entropy loss $\mathcal{L}_I$  weighted by 0.1. The classes correspond to 40 interaction categories on Inter-X, 26 caption-derived labels on InterHuman, and 37 conversation prompts on Seamless Interaction. At inference, each module uses a probability-weighted class embedding derived from its own predictions, without externally supplied labels.

\subsection{Placement module $P$}
\label{sec:P}

The placement module $P$ predicts the reactor's root
offset sequence from the actor's body and hand motion
$\mathbf{m}^{a}$.
Each output token represents two consecutive frames
and therefore contains six offset coordinates.
The module is conditioned on the reactor's initial
body and hand representation and its initial root offset:
\begin{equation}
    \mathbf{s}_P =
    [\mathbf{m}^{r}_1 \,\|\, \mathbf{o}_1].
\end{equation}
Because the body representation includes joint velocities computed by forward differences,
this initialization provides both the initial pose and velocity information derived from the first
two frames. The module uses a hidden dimension of $d=192$ and $N=6$ BiMamba blocks, and is sampled without
classifier-free guidance.

\subsection{Body module $B$}
\label{sec:B}

The body module $B$ generates reactor body motion conditioned on the actor body and a prescribed
root-offset sequence. Offsets are predicted by $P$ on Inter-X and InterHuman and fixed on Seamless Interaction.

\noindent\textbf{Placement-conditioned latent generation.}
Following ReactionMamba~\cite{beg2026reactionmamba}, we generate body motion in the latent space of a variational autoencoder~\cite{kingma2014vae}. The encoder produces 48-dimensional tokens
at half the motion frame rate, and the decoder additionally conditions on the actor body.
The VAE is pretrained and frozen during flow-model training.

Each flow token combines a body latent with the root offsets of its two corresponding frames:
\begin{equation}
    \mathbf{x}^{B}_1 =
    [\mathbf{z}\,\|\,\operatorname{patch}(\mathbf{o})]
    \in \mathbb{R}^{(T/2)\times54}.
\end{equation}
The six offset channels remain fixed throughout generation and are excluded from the flow-matching
loss. They contain ground-truth offsets during training and predicted offsets at inference.
Thus, $B$ generates body motion consistent with the placement supplied by $P$.

\noindent\textbf{Actor conditioning and initialization.}
Actor features are supplied at the network input and reinjected into every BiMamba block through
a learned gated residual connection: 
\begin{equation}
    \mathbf{h}'_l =
    \mathbf{h}_l
    + \omega_l W^a_l
    \left([\mathbf{h}_l\,\|\,\mathbf{K}]\right),
\end{equation}
where $W^a_l$ is a two-layer MLP and $\omega_l$ is
a learned gate.
This repeated conditioning is inspired by an encoder--decoder variant in the released
ReMFlow code~\cite{geng2026armflow}. The reactor seed contains the initial rotations of the 22 body joints and the first prescribed root offset.

\noindent\textbf{Dataset-specific conditioning.}
On Inter-X and InterHuman, sampling uses classifier-free guidance on actor motion and inferred intent.
On Seamless Interaction, both participants' speech features additionally condition $B$,
and sampling uses no guidance. Two auxiliary regularizers discourage insufficient arm movement and excessive
head motion during listening. Architecture settings, VAE objectives, guidance, and regularization details are
provided in the supplementary material.

\subsection{Hand module $H$}
\label{sec:H}

The hand module $H$ generates continuous 6D rotations for the reactor's 30 finger joints,
conditioned on the actor's body and hand motion and the reactor's body motion.
Its seed contains the observed initial reactor body and hand representation and the first offset
$\mathbf{o}_1$ (generated by $P$ at inference on Inter-X, fixed on Seamless Interaction): $\mathbf{s}_H = [\mathbf{m}^{r}_1\,\|\,\mathbf{o}_1]$.
The observed part of this seed is retained at inference, including the initial joint velocities
contained in $\mathbf{m}^{r}_1$. The body generated by $B$ enters only through the reactor-body conditioning stream.

On Seamless Interaction, each participant's speech features are additionally supplied to the corresponding motion stream.
A synchrony regularizer penalizes correlations between generated finger motion and reactor
speech energy that fall below the ground-truth reference. Architecture and loss details are provided in the supplementary material.
\subsection{Face module $F$}
\label{sec:F}

The face module $F$ generates 53-dimensional FLAME expression and jaw parameters on
Seamless Interaction. It conditions on the actor's body, hands, and masked facial parameters, together with
the reactor's body and both participants' speech features.

The seed combines the reactor's initial body, face, and first offset $\mathbf{o}_1$ (the fixed
placement on Seamless Interaction). At inference, its body component is taken from the output of $B$.
The facial component is set to zero whenever the first-frame face is unavailable or the reactor
is speaking at that frame, during both training and inference.

In addition to flow matching and intent classification, auxiliary losses supervise
temporal facial changes, jaw reconstruction, and jaw--speech synchrony.
Their definitions and implementation details are provided in the supplementary material; training and inference settings are given in the supplementary material.

\section{Experiments}
\label{sec:experiments}

\input{tabels/table1_interx}
\input{tabels/table2_interhuman}
\input{tabels/table3_interx_ablation}
\input{tabels/table4_seamless_body_hands}

\input{tabels/table6_seamless_face_transvae}

We describe datasets and metrics in Sec.~\ref{sec:datasets}, then compare ModularReactor with state-of-the-art methods on Inter-X, InterHuman, and Seamless Interaction (Sec.~\ref{sec:quantitative}).

\subsection{Datasets and metrics}
\label{sec:datasets}

\paragraph{Datasets.}
We evaluate on three datasets.
\textbf{Inter-X}~\cite{xu2024interx} provides
two-person SMPL-X body and hand motion,
interaction categories, and actor--reactor annotations.
\textbf{InterHuman}~\cite{liang2024intergen}
contains captioned two-person body motion
without articulated fingers.
For \textbf{Seamless Interaction}~\cite{agrawal2025seamless},
we construct a subset of conversations where there is clearly a main speaker (actor) and a listener (reactor)
using the UniLS-Talk index~\cite{chu2025unils},
with body, hand, facial, and speech features.
Dataset preparation and splits are detailed
in the supplementary material.

\noindent
\textbf{Physical interactions.}
On Inter-X and InterHuman, we use the released
text--motion evaluators through the ARMFlow
evaluation implementation~\cite{geng2026armflow}:
the Inter-X evaluator and InterGen's
InterCLIP evaluator~\cite{liang2024intergen},
respectively.
Following standard evaluation
practice~\cite{guo2022humanml3d}, we report
Fr\'echet inception distance (FID),
R-precision, multimodal distance (MM-Dist),
and diversity.
FID measures the discrepancy between generated
and real motion feature distributions.
R-precision measures text--motion retrieval
accuracy among the top-$k$ candidates, using candidate batches
of 32 on Inter-X and 96 on InterHuman.
MM-Dist measures the feature distance between
paired motions and captions.
Lower FID and MM-Dist and higher R-precision
are better; diversity is compared with
the ground-truth reference.

For methods evaluated by us, we additionally
report mean per-joint position error
(MPJPE, cm) and velocity error
(MPJVE, mm/frame) against the GT reaction.
On Inter-X, both errors are evaluated
on the same 400 actor-aligned test clips.

\noindent
\textbf{Conversational interactions.}
For body and hand motion, we report Fr\'echet gesture distance (FGD)~\cite{yoon2020speech}, following
the gesture-evaluation literature~\cite{kucherenko2023genea,nagy2026genea}.
FGD compares generated and real motion distributions in feature spaces learned
by autoencoders trained on our training windows.
We report FGD on the 22 body joints, the upper body (the 10 neck, head, and arm joints, scored with the GT lower body), and on the
30 finger joints (hands30).
For facial behavior, we adapt the REACT metrics~\cite{song2023react2023,song2025react} to FLAME parameters.
Facial reaction correlation (FRC) measures, for each generated sample, the highest
concordance correlation with a GT reference reaction associated with the same prompt.
Facial reaction distance (FRD) measures
the minimum dynamic-time-warping distance
to this reference set.
Higher FRC and lower FRD indicate better
agreement with the references.
We also report temporal variance (FRVar),
variation across generation seeds (S-MSE),
and variation across windows (FRDvs).
Finally, TLCC reports the absolute temporal
lag, in frames, at which cross-correlation
with the actor's facial motion is maximal.
Full metric definitions are provided in the supplementary material.

\subsection{Baselines}
\label{sec:baselines}

\noindent\textbf{Physical interactions.}
On Inter-X and InterHuman, we evaluate released ReMFlow~\cite{geng2026armflow} checkpoints.
We distinguish its official setting, which conditions on the first stored participant,
from a labelled-actor setting, which follows the annotated interaction order.
On InterHuman, we evaluate the released UMF~\cite{huang2026umf} checkpoint in an actor-given
adaptation: we encode the observed actor in place of its generated first participant and retain the
released reaction-generation stage, without retraining.

Caption-conditioned baselines receive the ground-truth interaction caption, whereas our model receives the reactor initialization
described in Section~\ref{sec:method}. The comparisons therefore involve different conditioning inputs, which are identified
in the tables.

\noindent\textbf{Conversational interactions.}
On Seamless Interaction, we retrain ConvoFusion~\cite{mughal2024convofusion} on our conversation windows for
body and hand generation, with its published inputs (the reactor's audio and both transcripts). We additionally evaluate a variant that also receives the reactor's first pose, as our chain does.
For facial reactions, we adapt the REACT Trans-VAE baseline~\cite{song2023react2023,song2025react}: a per-frame MLP on the partner's FLAME parameters replaces its video encoder, it receives the same speech features as our model and outputs FLAME parameters; the rest of the model is unchanged, retrained from scratch on our training windows.
All conversational baselines are evaluated on the same 276 test windows of the conversation-pair split (training details in the supplementary material).

\begin{figure*}[t]
\centering
\includegraphics[width=0.7\linewidth]{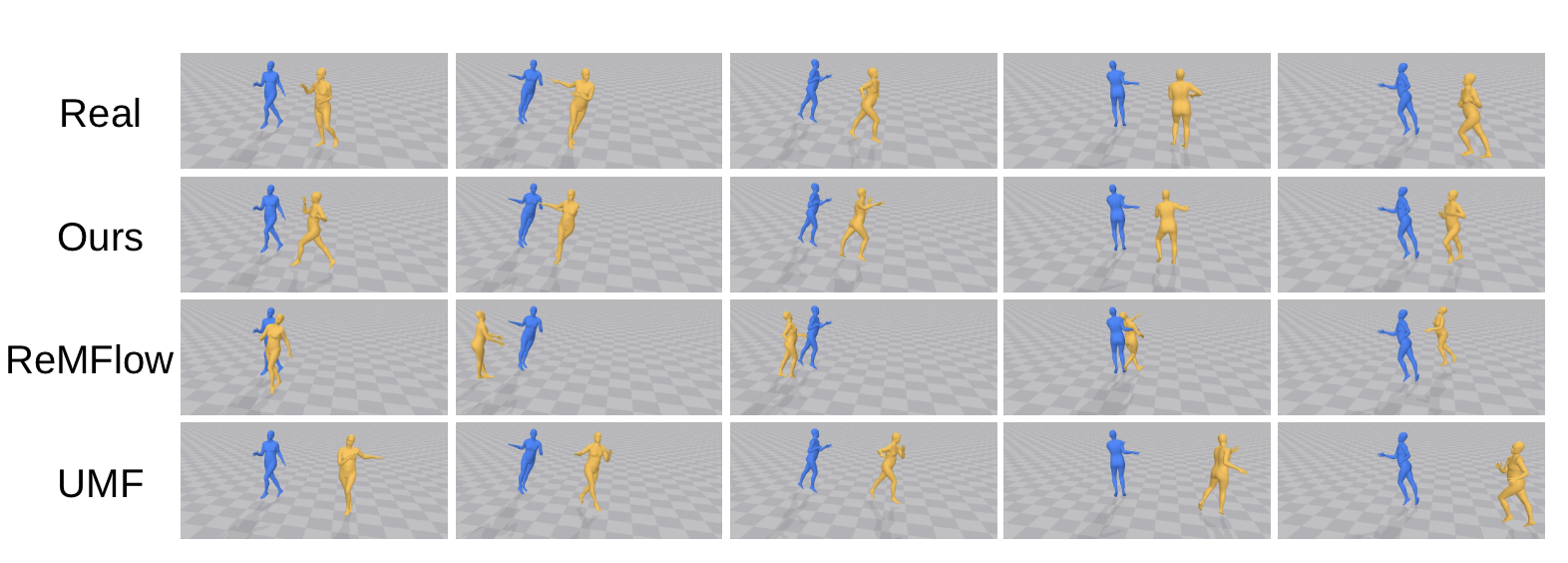}

\caption{Qualitative comparison on InterHuman sequence 2260
(dance) 
Rows: recorded actor {\color{blue} (blue)} with ground-truth, ours,
ReMFlow~\cite{geng2026armflow}, or UMF~\cite{huang2026umf}
reactor {\color{yellow} (yellow)}.
Columns share timestamps and camera views.
Baselines receive the caption and actor motion; ours receives reactor
initialization and actor motion.
Baseline renderings use output rotations; no finger motion is shown.}
\label{fig:qual}
\end{figure*}

\subsection{Quantitative results}
\label{sec:quantitative}

\noindent\textbf{Inter-X.}
Our text-free model achieves an FID comparable to
caption-conditioned ReMFlow (0.054 versus 0.061 for its rerun checkpoint and 0.058 as published;
overlapping confidence intervals, Table~\ref{tab:interx})
On the shared 400-clip subset, MPJPE and MPJVE decrease
from 42.2 to 27.2\,cm and from 20.03 to 12.93\,mm/frame,
respectively, while diversity remains close to GT.
Caption-conditioned methods retain better text--motion
alignment: ReMFlow achieves higher R@1
(0.440 versus 0.401) and lower MM-Dist.

\noindent\textbf{InterHuman.}
Our model obtains the lowest mean FID
(Table~\ref{tab:interhuman}), slightly below actor-given
UMF (1.656 versus 1.707), and reduces MPJPE and MPJVE
by approximately 20\% and 4\%.
However, R@1 remains lower (0.354 versus 0.398 for UMF
and 0.452 for published ReMFlow), and diversity is
farther from GT than for several baselines.

\noindent\textbf{Seamless Interaction.}
Table~\ref{tab:sl_BH} reports the FGD on the 276 test windows
(274 with valid FGD chunks). ConvoFusion is a co-speech gesture model driven by the reactor's audio and both transcripts, without the partner's motion; our chain receives the partner's motion and both audio tracks but no transcript. Compared with first-pose-conditioned ConvoFusion, our model lowers body22 FGD from 1.790 to 0.382 and upper FGD from 0.267 to 0.205.
Its hands30 FGD is higher (2.544 versus 2.133): finger gestures are
what ConvoFusion was designed for, with word-level timing from the
transcripts.

For faces (Table~\ref{tab:sl_face}), our full chain
achieves higher FRC than adapted Trans-VAE (0.148 versus 0.010), temporal variance close to GT faces (1.050 versus 0.985; Trans-VAE: 0.001), and higher diversity across seeds and windows (S-MSE 0.710 versus 0.024; FRDvs 2.814 versus 0.567).
Trans-VAE has lower FRD (32.72 versus 45.00), but holding
the first GT face constant already yields 35.23, so FRD
must be read together with temporal variation.
Using the GT body raises our FRC to 0.296, showing sensitivity to body-generation errors, and full-chain FRC remains only modestly above a GT face of another window (0.134): our faces are more dynamic than this Trans-VAE adaptation, but their correspondence with reference reactions remains limited.
Baseline training and checkpoint selection are detailed
in the supplementary material.

\subsection{Ablations}
\label{sec:ablations}
Table~\ref{tab:interx_ablation} (one component changed at a time; retrained variants use the same training recipe and duration) shows that removing the body seed nearly doubles FID (0.053 to 0.102). Replacing the VAE with PCA improves text alignment but increases FID and joint velocity error. A single network raises MPJPE from 27.1 to 33.5\,cm, supporting the modular hierarchy. Removing intent lowers caption-based alignment without increasing mean FID; guidance changes the scores little, with slightly lower joint errors without it. Seamless ablations are reported in Supplementary Material.

\subsection{Qualitative results}
\label{sec:qualitative}
Figure~\ref{fig:qual} shows a circular-dance clip from InterHuman (sequence 2260) included in the user study.
The real reactor circles with the actor at about one arm's length and turns its back to the camera in the fourth frame; our reactor keeps that distance, circles with the actor, and turns in the same frames. The ReMFlow reactor overlaps the actor in three of the five frames; the UMF reactor drifts farther from the actor.
Additional examples of body, hand, and facial reactions across the three datasets are provided in Supplementary Material and we will provide also videos animations.

\begin{table}[t]
\centering
\scriptsize
\setlength{\tabcolsep}{3pt}
\caption{User study results (21 raters).
Percentage of preferences for our method versus each competing method, evaluated on reaction appropriateness and motion naturalness.
Win: our method is preferred; Tie: no clear preference; Loss: the competing method is preferred.
Each rater viewed each clip once, with presentation sides and pairings balanced.}
\label{tab:user_study}
\begin{tabular}{@{}llrrrrrr@{}}
\toprule
& & \multicolumn{3}{c}{Appropriateness}
  & \multicolumn{3}{c}{Naturalness} \\
\cmidrule(lr){3-5} \cmidrule(lr){6-8}
Dataset & Ours vs. & Win & Tie & Loss & Win & Tie & Loss \\
\midrule
Inter-X & ReMFlow & 57\% & 19\% & 24\% & 59\% & 22\% & 19\% \\
& GT & 22\% & 17\% & 60\% & 19\% & 14\% & 67\% \\
\midrule
InterHuman & ReMFlow & 45\% & 31\% & 24\% & 43\% & 21\% & 36\% \\
& UMF & 33\% & 21\% & 45\% & 38\% & 24\% & 38\% \\
& GT & 2\% & 14\% & 83\% & 2\% & 17\% & 81\% \\
\midrule
Seamless & ConvoFusion (first pose) & 27\% & 29\% & 44\% & 30\% & 27\% & 43\% \\
& GT & 14\% & 16\% & 70\% & 13\% & 14\% & 73\% \\
\bottomrule
\end{tabular}
\end{table}
\subsection{User Study}
\label{sec:user_study}
We conducted a blinded A/B comparison study with 21 raters to assess reaction appropriateness and motion naturalness. The study covered physical interactions (Inter-X and
InterHuman) and conversational body and hand motion (Seamless Interaction).
Six test clips per dataset were selected from GT-motion previews before rendering model outputs.
Comparators were ReMFlow, additionally UMF on InterHuman, and ConvoFusion given the first pose for conversational body and hands, with GT reactions as references.
Every rater saw each clip once, with comparisons and sides balanced across raters.
Table~\ref{tab:user_study} reports the share of judgements for ours, for ``no clear difference'', and for the other video.
More details are given in the supplementary material.

\noindent\textbf{Results.}
On Inter-X, raters favor our model over caption-conditioned ReMFlow in appropriateness (57\% against 24\%) and naturalness (59\% against 19\%).
On InterHuman, they favor ours over ReMFlow (45\% against 24\%, 43\% against 36\%) but UMF in appropriateness (45\% against 33\%), with a tie in naturalness (38\% each).
On Seamless Interaction, ConvoFusion is chosen more often (44\% against 27\%, 43\% against 30\%, with 29\% and 27\% ties): it is a model built for co-speech gestures, whereas our general
reaction chain is still judged at least as good in 56\% and 57\% of the cases once ties are counted. GT reactions are preferred throughout, most clearly on InterHuman
(83\% and 81\%) and Seamless Interaction (70\% and 73\%).

\section{Conclusion and limitations}

We introduced ModularReactor, an offline framework
for physical and conversational reaction generation
without captions or action labels at inference.
Its hierarchy $P\rightarrow B\rightarrow\{H,F\}$
combines independently trained flow-matching modules
with inferred intent embeddings.
Results on Inter-X and InterHuman show competitive
FID and lower joint errors, while caption-conditioned
baselines retain stronger text--motion alignment.
On Seamless Interaction, our model improves body FGD
over a retrained ConvoFusion but not hand FGD.
It generates more temporally varied facial behavior
than our adapted Trans-VAE baseline.
Ablations support the modular decomposition.

\noindent\textbf{Limitations.}
Our framework requires the complete actor sequence
and a reactor initialization, and does not explicitly
enforce contact or prevent interpenetration.
Modules are trained separately per dataset,
and errors can propagate through the hierarchy,
particularly from the body to the face.
Conversational generation assumes fixed placement
and recorded speech from both participants.
Moreover, automatic metric gains do not consistently
translate into demonstrated perceptual improvements,
and the facial comparison is limited by the
Trans-VAE adaptation.
Future work will address contact-aware generation,
robustness to upstream errors, and online synthesis
with jointly generated motion and speech.

\section*{Acknowledgements}
This project is supported by the France 2030 program (ANR-22-EXEN-0004, PEPR eNSEMBLE / MATCHING).
\section*{LLM Usage Statement}

During the preparation of this work, the authors used ChatGPT (OpenAI) and Claude (Anthropic) as an assistant for three purposes: (i) editing the manuscript text to improve clarity, grammar, and readability; (ii) writing, refining and debugging parts of the source code; and (iii) helping produce the figures and supplementary videos, with the aim of presenting high-quality visualizations. All visual results shown in the figures and videos are produced by the authors' method and the compared baselines; the LLM was used only to improve their presentation, not to generate or alter the underlying results. The research ideas, method design, experimental setup, analysis of results, and conclusions are entirely the work of the authors. The authors reviewed and verified all AI-assisted text, code, and visual material, and take full responsibility for the content of this article.

%% file: tabels/table1_interx.tex
\begin{table*}[!t]
\centering
\scriptsize
\setlength{\tabcolsep}{4pt}

\caption{Reaction generation on Inter-X.
[P]: published~\cite{geng2026armflow};
[R]: checkpoint reruns; [O]: trained by us.
Baselines receive captions; ours receives reactor initialization.
Mean $\pm$ 95\% CI over 20 seeds.
MPJPE (cm) and MPJVE (mm/frame): 400 actor-aligned clips.
Bold/underline: best/second best, excluding GT;
Diversity: closest to GT (8.93) in bold.}
\label{tab:interx}

\resizebox{\textwidth}{!}{%
\begin{tabular}{@{}llccccccc@{}}
\toprule
Method & Prov.
& FID$\downarrow$
& R@1$\uparrow$
& R@3$\uparrow$
& MM-Dist$\downarrow$
& Div.\ ($\to$GT)
& MPJPE$\downarrow$
& MPJVE$\downarrow$ \\
\midrule

\emph{Ground truth} & [O]
& \emph{0.005$\pm$.000}
& \emph{0.434$\pm$.004}
& \emph{0.740$\pm$.003}
& \emph{3.590$\pm$.001}
& \emph{8.934$\pm$.046}
& \emph{0}
& \emph{0} \\
\midrule

InterGen & [P]
& 0.238$\pm$.038
& 0.352$\pm$.004
& 0.643$\pm$.004
& 4.212$\pm$.029
& 8.773$\pm$.067
& ---
& --- \\

InterMask & [P]
& 0.399$\pm$.013
& \underline{0.429}$\pm$.005
& \underline{0.731}$\pm$.005
& \underline{3.584}$\pm$.017
& 8.911$\pm$.057
& ---
& --- \\

MLD & [P]
& 0.148$\pm$.020
& 0.414$\pm$.004
& 0.712$\pm$.004
& 3.655$\pm$.020
& 8.893$\pm$.053
& ---
& --- \\

ReGenNet & [P]
& 0.093$\pm$.022
& 0.407$\pm$.004
& 0.705$\pm$.003
& 3.762$\pm$.015
& 8.841$\pm$.072
& ---
& --- \\

ReMFlow & [P]
& \underline{0.058}$\pm$.005
& \textbf{0.440}$\pm$.004
& \textbf{0.743}$\pm$.003
& \textbf{3.570}$\pm$.013
& \textbf{8.948}$\pm$.068
& ---
& --- \\

ReMFlow, official setting & [R]
& 0.061$\pm$.003
& 0.415$\pm$.003
& 0.709$\pm$.004
& 3.751$\pm$.017
& 8.999$\pm$.067
& \underline{42.2}$\pm$.1
& \underline{20.03}$\pm$.04 \\

ReMFlow, labelled actor & [R]
& 0.152$\pm$.003
& 0.410$\pm$.004
& 0.702$\pm$.002
& 3.760$\pm$.008
& 8.906$\pm$.076
& \underline{42.2}$\pm$.1
& \underline{20.03}$\pm$.04 \\

\rowcolor{gray!15}
\textbf{Ours (no caption)} & [O]
& \textbf{0.054}$\pm$.003
& 0.401$\pm$.004
& 0.693$\pm$.004
& 3.861$\pm$.008
& 8.902$\pm$.044
& \textbf{27.2}$\pm$.2
& \textbf{12.93}$\pm$.04 \\

\bottomrule
\end{tabular}%
}

\end{table*}


%% file: tabels/table2_interhuman.tex
\begin{table*}[!t]
\centering
\scriptsize
\setlength{\tabcolsep}{4pt}

\caption{Reaction generation on InterHuman.
Our text-free model achieves the lowest FID and joint errors,
while caption-conditioned baselines retain higher R-precision.
[P], [R], [O]: as in Table~\ref{tab:interx}.
Baselines receive ground-truth captions; ours receives no caption.
ARMFlow evaluator; mean $\pm$ 95\% CI over 20 seeds.
MPJPE (cm) and MPJVE (mm/frame): clips with at least 40 frames.
Diversity is compared with ground truth (7.74), without ranking.
Bold / underline: best / second best, excluding ground truth.}
\label{tab:interhuman}

\resizebox{\textwidth}{!}{%
\begin{tabular}{@{}llcccccccc@{}}
\toprule
Method & Prov.
& FID$\downarrow$
& R@1$\uparrow$
& R@2$\uparrow$
& R@3$\uparrow$
& MM-Dist$\downarrow$
& Div.\ ($\to$GT)
& MPJPE$\downarrow$
& MPJVE$\downarrow$ \\
\midrule

\emph{Ground truth} & [O]
& \emph{0.357$\pm$.005}
& \emph{0.420$\pm$.004}
& \emph{0.599$\pm$.003}
& \emph{0.695$\pm$.004}
& \emph{3.786$\pm$.000}
& \emph{7.744$\pm$.038}
& \emph{0}
& \emph{0} \\
\midrule

InterGen & [P]
& 9.183$\pm$.174
& 0.325$\pm$.004
& 0.467$\pm$.004
& 0.546$\pm$.005
& 3.859$\pm$.001
& 7.305$\pm$.047
& --- & --- \\

in2IN & [P]
& 7.913$\pm$.251
& 0.362$\pm$.004
& 0.504$\pm$.008
& 0.589$\pm$.007
& 3.832$\pm$.002
& 7.709$\pm$.040
& --- & --- \\

MLD & [P]
& 3.588$\pm$.076
& 0.405$\pm$.005
& 0.561$\pm$.007
& 0.649$\pm$.006
& 3.798$\pm$.002
& 7.663$\pm$.040
& --- & --- \\

ReGenNet & [P]
& 2.930$\pm$.052
& 0.362$\pm$.005
& 0.513$\pm$.005
& 0.605$\pm$.004
& 3.815$\pm$.001
& 7.582$\pm$.064
& --- & --- \\

InterMask & [P]
& 3.453$\pm$.061
& \underline{0.451}$\pm$.007
& \underline{0.610}$\pm$.006
& \underline{0.701}$\pm$.005
& \underline{3.782}$\pm$.002
& 7.710$\pm$.046
& --- & --- \\

ReMFlow & [P]
& 2.433$\pm$.042
& \textbf{0.452}$\pm$.006
& \textbf{0.618}$\pm$.005
& \textbf{0.708}$\pm$.005
& \textbf{3.778}$\pm$.001
& 7.714$\pm$.029
& --- & --- \\

ReMFlow, official setting & [R]
& 2.476$\pm$.021
& 0.432$\pm$.004
& 0.599$\pm$.004
& 0.691$\pm$.003
& 3.785$\pm$.000
& 7.715$\pm$.026
& 67.1$\pm$.3
& 32.25$\pm$.08 \\

ReMFlow, labelled actor & [R]
& 2.604$\pm$.018
& 0.435$\pm$.005
& 0.599$\pm$.004
& 0.690$\pm$.002
& 3.783$\pm$.000
& 7.706$\pm$.025
& 68.0$\pm$.2
& 32.23$\pm$.07 \\

UMF, react mode & [R]
& \underline{1.707}$\pm$.020
& 0.398$\pm$.003
& 0.565$\pm$.004
& 0.657$\pm$.003
& 3.797$\pm$.000
& 7.830$\pm$.034
& \underline{32.8}$\pm$.2
& \underline{20.25}$\pm$.04 \\

\rowcolor{gray!15}
\textbf{Ours} (no caption) & [O]
& \textbf{1.656}$\pm$.021
& 0.354$\pm$.005
& 0.509$\pm$.003
& 0.602$\pm$.004
& 3.818$\pm$.000
& 7.858$\pm$.038
& \textbf{26.2}$\pm$.1
& \textbf{19.36}$\pm$.04 \\

\bottomrule
\end{tabular}%
}
\end{table*}

%% file: tabels/table3_interx_ablation.tex
\begin{table*}[!t]
\centering
\scriptsize
\setlength{\tabcolsep}{4pt}

\caption{Ablation study on Inter-X, changing one component
at a time from the final model.
All variants use the same GPU, evaluation protocol, and
10 generation seeds (mean $\pm$ 95\% confidence interval).
Differences from Table~\ref{tab:interx} reflect the seed
count and GPU.
Bold: best, excluding ground truth.
Diversity is compared with GT (8.92), without ranking.
MPJPE (cm) and MPJVE (mm/frame) follow
Table~\ref{tab:interx}.}
\label{tab:interx_ablation}

\resizebox{\textwidth}{!}{%
\begin{tabular}{@{}p{6.4cm}ccccccc@{}}
\toprule
Model
& FID$\downarrow$
& R@1$\uparrow$
& R@3$\uparrow$
& MM-Dist$\downarrow$
& Div.\ ($\to$GT)
& MPJPE$\downarrow$
& MPJVE$\downarrow$ \\
\midrule

Ground truth (same conversion and seeds)
& 0.005$\pm$.001
& 0.439$\pm$.004
& 0.742$\pm$.004
& 3.590$\pm$.002
& 8.917$\pm$.066
& 0
& 0 \\
\midrule

\rowcolor{gray!15}
\textbf{Final model } 
($P\rightarrow B\rightarrow H$, latent $B$,
first-frame seed, guidance)
& 0.053$\pm$.003
& 0.398$\pm$.005
& 0.693$\pm$.006
& 3.874$\pm$.014
& 8.892$\pm$.072
& 27.1$\pm$.3
& 12.94$\pm$.05 \\
\midrule

w/o guidance ($w_a=w_s=1$; sampling only)
& 0.057$\pm$.004
& 0.393$\pm$.006
& 0.691$\pm$.006
& 3.891$\pm$.014
& 8.833$\pm$.073
& \textbf{26.8}$\pm$.3
& 12.58$\pm$.05 \\

$B$ w/o first-frame seed (retrained)
& 0.102$\pm$.008
& 0.378$\pm$.005
& 0.672$\pm$.007
& 4.003$\pm$.012
& 8.870$\pm$.085
& 28.7$\pm$.2
& 13.05$\pm$.06 \\

$B$ w/o VAE: PCA tokens, $k=250$ (retrained)
& 0.071$\pm$.006
& \textbf{0.408}$\pm$.003
& \textbf{0.702}$\pm$.005
& \textbf{3.825}$\pm$.011
& 8.906$\pm$.073
& 28.0$\pm$.2
& 16.70$\pm$.05 \\

$P$, $B$, and $H$ w/o intent (retrained)
& \textbf{0.049}$\pm$.004
& 0.378$\pm$.008
& 0.668$\pm$.007
& 4.024$\pm$.018
& 8.756$\pm$.068
& 28.2$\pm$.3
& \textbf{12.54}$\pm$.07 \\

Single network for root, body, and fingers; (445-d VAE)
& 0.069$\pm$.003
& 0.400$\pm$.005
& 0.698$\pm$.006
& 3.827$\pm$.012
& 8.888$\pm$.076
& 33.5$\pm$.2
& 18.42$\pm$.06 \\

\bottomrule
\end{tabular}%
}
\end{table*}

%% file: tabels/table4_seamless_body_hands.tex

\begin{table}[t]
\centering
\scriptsize
\setlength{\tabcolsep}{3pt}

\caption{Body and hand generation on Seamless Interaction. 276 test windows; mean and 95\% CI over 20 runs (one seed per run).
ConvoFusion~\cite{mughal2024convofusion} is retrained on our windows with its published inputs (reactor audio, both transcripts); ``first pose'' also receives the reactor's first pose, as our chain does. Placement is fixed in every row. Bold: best.}

\label{tab:sl_BH}

\resizebox{\columnwidth}{!}{%
\begin{tabular}{lccc}
\toprule
& Body22 FGD$\downarrow$
& Upper FGD$\downarrow$
& Hands30 FGD$\downarrow$ \\
\midrule







Ours
& \textbf{0.382}{\scriptsize$\pm$.015}
& \textbf{0.205}{\scriptsize$\pm$.004}
& 2.544{\scriptsize$\pm$.047} \\

ConvoFusion~\cite{mughal2024convofusion}
& 2.674{\scriptsize$\pm$.067}
& 0.291{\scriptsize$\pm$.005}
& 2.208{\scriptsize$\pm$.061} \\

ConvoFusion, first pose
& 1.790{\scriptsize$\pm$.053}
& 0.267{\scriptsize$\pm$.005}
& \textbf{2.133}{\scriptsize$\pm$.060} \\

\bottomrule
\end{tabular}%
}

\end{table}

%% file: tabels/table6_seamless_face_transvae.tex
\begin{table}[t]
\centering
\scriptsize
\setlength{\tabcolsep}{2pt}

\caption{Facial reactions on Seamless Interaction
(276 windows): our full chain versus adapted
Trans-VAE~\cite{song2023react2023,song2025react}.
References: GT face, and our face module conditioned on the GT body.
Mean $\pm$ 95\% CI over 20 runs. Bold: best of the two generated rows.}

\label{tab:sl_face}

\resizebox{\columnwidth}{!}{%
\begin{tabular}{@{}lcccccc@{}}
\toprule
& \shortstack{FRC/seed\\$\uparrow$}
& \shortstack{FRD\\$\downarrow$}
& \shortstack{FRVar\\($\to$0.985)}
& \shortstack{S-MSE\\$\uparrow$}
& \shortstack{FRDvs\\$\uparrow$}
& \shortstack{TLCC\\$\downarrow$} \\
\midrule

\multicolumn{7}{l}{\emph{References}} \\

GT face & 1.000 & 0.00 & 0.985 & -- & 4.507 & 35.5 \\


Ours, GT body
& 0.296{\scriptsize$\pm$.000}
& 40.83{\scriptsize$\pm$.01}
& 1.021{\scriptsize$\pm$.001}
& 0.643{\scriptsize$\pm$.001}
& 2.835{\scriptsize$\pm$.002}
& 32.5{\scriptsize$\pm$.1} \\

\midrule

\multicolumn{7}{l}{\emph{20 runs}} \\

Ours, full chain
& \textbf{0.148}{\scriptsize$\pm$.001}
& 45.00{\scriptsize$\pm$.08}
& \textbf{1.050}{\scriptsize$\pm$.006}
& \textbf{0.710}{\scriptsize$\pm$.002}
& \textbf{2.814}{\scriptsize$\pm$.023}
& 32.7{\scriptsize$\pm$.3} \\

Trans-VAE
& 0.010{\scriptsize$\pm$.000}
& \textbf{32.72}{\scriptsize$\pm$.01}
& 0.001{\scriptsize$\pm$.000}
& 0.024{\scriptsize$\pm$.000}
& 0.567{\scriptsize$\pm$.001}
& \textbf{32.5}{\scriptsize$\pm$.1} \\

\bottomrule
\end{tabular}%
}

\end{table}

%% file: sec/X_suppl.tex
\clearpage
\maketitlesupplementary

\appendix
\setcounter{table}{0}
\setcounter{figure}{0}
\setcounter{equation}{0}
\renewcommand{\thetable}{S\arabic{table}}
\renewcommand{\thefigure}{S\arabic{figure}}
\renewcommand{\theequation}{S\arabic{equation}}

This supplementary material extends the main paper with an ethical statement (Section~\ref{sec:supp_ethics}), the related work in full (Section~\ref{sec:supp_related}), the motion and speech representations (Section~\ref{sec:supp_rep}), the flow parameterization and sampler (Section~\ref{sec:supp_fm}), the architecture, objectives, and training settings of the four modules (Section~\ref{sec:supp_impl}), the datasets and splits (Section~\ref{sec:supp_datasets}), the conversational baselines (Section~\ref{sec:supp_baselines}), the evaluation metrics (Section~\ref{sec:metrics}), the Seamless Interaction ablations (Section~\ref{sec:supp_sl_ablations}), the user studies (Section~\ref{sec:supp_user_study}), and additional qualitative results (Section~\ref{sec:supp_qual}).
The supplementary videos show the clips of the user studies for every compared method.

\section{Ethical statement}
\label{sec:supp_ethics}
ModularReactor is intended for beneficial applications in virtual reality, digital human modeling and accessibility tools. However, as for any technology capable of synthesizing realistic human motion, we acknowledge the risk of misuse. In particular, we strongly condemn impersonation, deceptive synthetic media, and any application that animates the representation of an individual without their informed and explicit consent.

Regarding data, all experiments rely on publicly available datasets released for research with participant consent; no new motion or video data was collected, and no personally identifiable information beyond what is present in the original published datasets is used or disclosed.

After the review process, we commit to releasing code and pretrained weights under a license that explicitly prohibits non-consensual animation of real individuals.

\section{Extended related work}
\label{sec:supp_related}
This section extends the related work of Section~\mainSecRelated{} of the main paper. Table~\ref{tab:positioning_full} compares the conditioning inputs, reactor initialization, motion representation, and generated outputs of the methods discussed there.

\begin{table*}[t]
\centering
\scriptsize
\setlength{\tabcolsep}{3pt}
\renewcommand{\arraystretch}{1.12}

\caption{Comparison of inference inputs, reactor initialization,
motion representation, and generated outputs.
``Actor'' denotes the observed interaction partner.
Parentheses indicate optional inputs or extensions.
Our framework generates dataset-dependent components:
placement on Inter-X and InterHuman, hands on Inter-X
and Seamless Interaction, and face on Seamless Interaction.}
\label{tab:positioning_full}

\begin{tabularx}{\textwidth}{@{}
>{\raggedright\arraybackslash}p{0.15\textwidth}
>{\raggedright\arraybackslash}X
>{\raggedright\arraybackslash}p{0.12\textwidth}
>{\raggedright\arraybackslash}p{0.21\textwidth}
>{\raggedright\arraybackslash}p{0.18\textwidth}
@{}}
\toprule
Method
& Conditioning at inference
& Reactor initialization
& Motion representation
& Generated output \\
\midrule

\multicolumn{5}{@{}l}{
\textit{Text-conditioned interaction generation}} \\
\addlinespace[2pt]

Interact2Ar~\cite{ruizponce2026interact2ar}
& Interaction caption
& ---
& SMPL-X rotations (6D) + trajectory
& Both participants: trajectory, body, hands \\

\midrule
\multicolumn{5}{@{}l}{
\textit{Motion-conditioned reaction generation}} \\
\addlinespace[2pt]

InterFormer~\cite{chopin2023interformer}
& Actor motion
& Initial pose
& Joint positions (Kinect skeletons)
& Body \\

ReGenNet~\cite{xu2024regennet}
& Actor motion (+ action label)
& ---
& SMPL-X rotations + translation
& Body, hands \\

ReMoS~\cite{ghosh2024remos}
& Actor motion
& ---
& Joint positions
& Body, then hands \\

Think-Then-React~\cite{tan2025ttr}
& Actor motion; text generated internally
& ---
& InterGen features (positions, velocities, rotations)
& Body \\

ReMFlow~\cite{geng2026armflow}
& Actor motion, interaction caption
& ---
& SMPL-X rotations on Inter-X; InterGen features on InterHuman
& Body; hands on Inter-X \\

UMF~\cite{huang2026umf}, reaction mode
& Actor motion, interaction caption
& ---
& InterGen features
& Body \\

ReactionMamba~\cite{beg2026reactionmamba}
& Actor motion
& First frame
& Joint positions
& Body, hands \\

MoReact~\cite{xu2025moreact}
& Actor motion, caption or action label
& ---
& HumanML3D-style features (positions, velocities, rotations, contacts)
& Trajectory, then body \\

Ready-to-React~\cite{cen2025readytoreact}
& Motion history of both participants
& First 4 frames
& SMPL body rotations
& Body (online) \\

ReMoGen~\cite{ye2026remogen}
& Actor motion history (+ text intent, scene)
& First 2 frames
& SMPL-X-based features (6D rotations, positions, velocities), latent prior
& Body (online) \\

\midrule
\multicolumn{5}{@{}l}{
\textit{Conversational motion and facial reaction generation}} \\
\addlinespace[2pt]

ConvoFusion~\cite{mughal2024convofusion}
& Own speech audio and transcript, partner's transcript
& ---
& Joint positions
& Body, hands \\

DyaDiT~\cite{peng2026dyadit}
& Audio of both participants, own reference motion
(+ partner motion, social attributes)
& 16-frame prefix (configurable)
& SMPL-H rotations (6D), 43 upper-body joints
& Upper body, fingers \\

DyaPlex~\cite{nagano2026dyaplex}
& Partner speech and motion; fixed system prompt
& ---
& SMPL-X pose parameters
& Speech, body, hands (+ face in a demo variant) \\

Audio2Photoreal~\cite{ng2024audio2photoreal}
& Audio of both participants
& ---
& Joint angles; face codes
& Body, hands, face \\

ReactMotion~\cite{luo2026reactmotion}
& Speaker text, audio, or both (+ emotion)
& ---
& HumanML3D features
& Listener body \\

UniLS~\cite{chu2025unils}
& Audio of both participants
& ---
& FLAME expression, jaw, head pose
& Face, head \\

Trans-VAE~\cite{song2023react2023}
& Partner face video and audio
& ---
& 3DMM coefficients + facial attributes
& Face \\

\midrule
\rowcolor{green!15}
\textbf{ModularReactor (ours)}
& \textbf{Actor motion (+ audio of both participants)}
& \textbf{Initial pose and joint velocities}
& \textbf{ARDY body features (positions, 6D rotations, velocities, contacts); 6D fingers; root offsets; FLAME face}
& \textbf{Placement, body, hands, face} \\

\bottomrule
\end{tabularx}

\smallskip
\parbox{\textwidth}{\scriptsize
``---'' means that initialization is not specified
in this comparison, rather than necessarily absent.
Our initialization is module-specific:
joint velocities, where used, are computed from
the first two reactor frames.
Our method requires no caption or externally supplied
action or intent label at inference.
}
\end{table*}

\section{Motion representations and audio preprocessing}
\label{sec:supp_rep}

\noindent\textbf{Placement and body.}
Both participants are expressed in a common coordinate system whose origin is the actor's root position in the first frame. The reactor's placement is the root offset $\mathbf{o}_i=\mathbf{r}^{r}_i-\mathbf{r}^{a}_i$.
Following ARDY~\cite{zhao2026ardy}, the body representation is
\begin{equation}
\mathbf{b}_i = [\mathbf{p}_i\,\|\,\mathbf{R}_i\,\|\,
\dot{\mathbf{j}}_i\,\|\,\boldsymbol{\kappa}_i]\in\mathbb{R}^{265},
\end{equation}
with the pelvis-relative positions of 21 joints $\mathbf{p}_i\in\mathbb{R}^{63}$, the continuous 6D
rotations~\cite{zhou2019continuity} of the 22 SMPL-X body joints~\cite{pavlakos2019smplx} $\mathbf{R}_i\in\mathbb{R}^{132}$, the world-space velocities of the 22 joints $\dot{\mathbf{j}}_i\in\mathbb{R}^{66}$, and four binary foot-contact indicators $\boldsymbol{\kappa}_i$ for the ankle and toe joints of both feet, as in HumanML3D~\cite{guo2022humanml3d}.

\noindent\textbf{Hands and face.}
The 30 finger joints use 6D rotations relative to their parent joint, $\mathbf{h}_i\in\mathbb{R}^{180}$, and body and hands together give
$\mathbf{m}_i=[\mathbf{b}_i\,\|\,\mathbf{h}_i]\in\mathbb{R}^{445}$. On Seamless Interaction, the face is represented by 50 FLAME expression coefficients~\cite{li2017flame} and a three-dimensional jaw rotation,
$\mathbf{f}_i\in\mathbb{R}^{53}$, with a binary validity flag per frame.

\noindent\textbf{Speech.}
Each participant's audio is encoded with wav2vec2-large-robust fine-tuned on Switchboard~\cite{baevski2020wav2vec}.
The 1,024-dimensional features of its last hidden layer are averaged to the 25\,fps motion rate and projected on 64 principal components, $\boldsymbol{\alpha}_i\in\mathbb{R}^{64}$. The projection and the scaling are fitted on the training windows only.

\section{Flow parameterization and sampling}
\label{sec:supp_fm}

\noindent\textbf{Recovering the sample and velocity.}
Let $\mathbf{x}_0$ be Gaussian noise and $\mathbf{x}_1$ the real motion, with the interpolation $\mathbf{x}_t=(1-t)\mathbf{x}_0+t\mathbf{x}_1$.
Given the conditioning $\mathbf{c}$ (Section~\ref{sec:supp_network}), the network predicts $\hat{\mathbf{v}}_t=g_\theta(\mathbf{x}_t,t,\mathbf{c})$, whose target is $\mathbf{v}_t=t\mathbf{x}_0-(1-t)\mathbf{x}_1$.
This target differs from the interpolation velocity $\mathbf{u}_t=\mathbf{x}_1-\mathbf{x}_0$ used for sampling.
Solving these equations for the clean sample and the velocity gives
\begin{equation}
\begin{aligned}
\hat{\mathbf{x}}_1
&=
\frac{
  t\mathbf{x}_t-(1-t)\hat{\mathbf{v}}_t
}{
  t^2+(1-t)^2
},\\
\hat{\mathbf{u}}_t
&=
\frac{
  (2t-1)\mathbf{x}_t-\hat{\mathbf{v}}_t
}{
  t^2+(1-t)^2
}.
\end{aligned}
\end{equation}
The clean-sample estimate is also used to evaluate the auxiliary objectives of the hand and face modules.

\noindent\textbf{Numerical sampling.}
Sampling starts from Gaussian noise and uses the time grid $t_k=k/25$, $k=0,\ldots,24$,
with an explicit Euler update $\mathbf{x}_{t_{k+1}}=\mathbf{x}_{t_k}+\frac{1}{25}\hat{\mathbf{u}}_{t_k}$
for $k<24$; at $t_{24}$ we return $\hat{\mathbf{x}}_1$ instead of a last Euler step.
For $B$, the prescribed offset channels are held fixed as described in Section~\ref{sec:supp_B}.

\section{Implementation and training details}
\label{sec:supp_impl}

\subsection{Shared network}
\label{sec:supp_network}

\begin{figure*}[t]
\centering
\includegraphics[width=\textwidth]{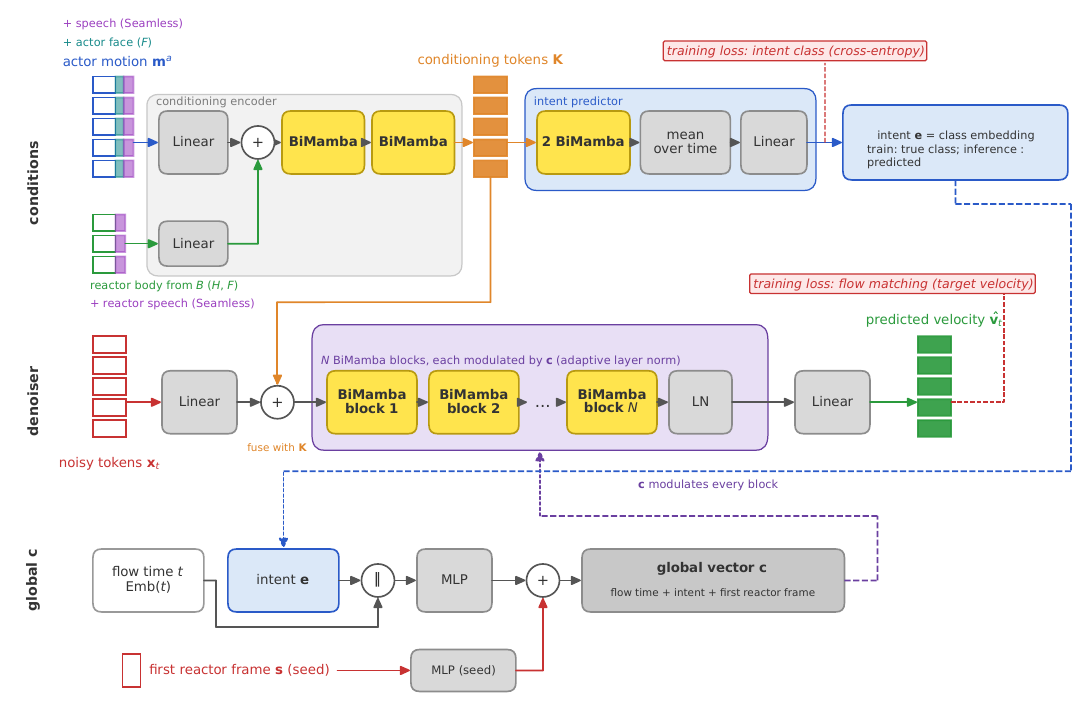}
\caption{The network shared by the four modules. A conditioning encoder turns the conditioning streams (the actor stream and, for $H$ and $F$, the reactor body generated by $B$; on Seamless Interaction, the speech
features of both participants) into conditioning tokens $\mathbf{K}$.
An intent predictor reads $\mathbf{K}$ and gives the intent $\mathbf{e}$. The flow time, the intent, and the reactor seed form the global vector $\mathbf{c}$. A denoiser of $N$ bidirectional Mamba blocks, modulated by $\mathbf{c}$, predicts $\hat{\mathbf{v}}_t$.}
\label{fig:shared_network}
\end{figure*}

Figure~\ref{fig:shared_network} shows the network shared by the four modules.

\noindent\textbf{Conditioning and intent prediction.}
Each conditioning stream is mapped to a common $d$-dimensional space by a linear projection.
The projected streams are summed and processed by two BiMamba blocks, giving the conditioning tokens $\mathbf{K}\in\mathbb{R}^{(T/2)\times d}$. 
An intent predictor processes $\mathbf{K}$ with two further BiMamba blocks, temporal mean pooling, and a linear head that outputs $K$ logits $\boldsymbol{\ell}$, supervised by a cross-entropy loss $\mathcal{L}_I$.
A learned table $\mathbf{E}$ maps classes to embeddings. At inference, every module uses its own probability-weighted embedding $\mathbf{e}=\operatorname{softmax}(\boldsymbol{\ell})^\top\mathbf{E}_{1:K}$.
During training, the embedding of the GT class is used; $B$ uses the predicted embedding in 25\% of the samples.
On InterHuman, clips without a label are excluded from $\mathcal{L}_I$ and use the predicted embedding.

\noindent\textbf{Denoising network.}
A global vector combines the flow-time embedding, the intent embedding, and the reactor seed:
$\mathbf{c}=\operatorname{MLP}([\operatorname{Emb}(t)\,\|\,\mathbf{e}])
+\operatorname{MLP}^{\mathrm{seed}}(\mathbf{s})$.
The seed enters only through $\mathbf{c}$, so every frame is generated, including the first.
The noisy motion tokens are concatenated with the conditioning tokens $\mathbf{K}$, as in the actor--reactor fusion of
ARMFlow~\cite{geng2026armflow}.
They are processed by $N$ BiMamba blocks modulated by zero-initialized adaptive layer normalization~\cite{peebles2023dit}:
\begin{equation}
\begin{aligned}
\tilde{\mathbf{h}}_l &= \operatorname{LN}(\mathbf{h}_l)\odot
(1+\boldsymbol{\gamma}_l)+\boldsymbol{\beta}_l,\\
\mathbf{h}_{l+1} &= \mathbf{h}_l+\mathbf{g}_l\odot
\operatorname{BiMamba}_l(\tilde{\mathbf{h}}_l),\\
[\boldsymbol{\beta}_l,\boldsymbol{\gamma}_l,\mathbf{g}_l] &= W_l\mathbf{c}.
\end{aligned}
\end{equation}
During training, a learned null token replaces $\mathbf{K}$ in 15\% of the samples.
In $B$, whose guidance acts on both $\mathbf{K}$ and the intent, the samples are split into four exclusive cases: 75\% keep both conditions, 10\% lose only $\mathbf{K}$, 10\% lose only the intent, and 5\% lose both.
Each condition is therefore dropped in 15\% of the samples ($10\%+5\%$).

\subsection{Placement module $P$}
\label{sec:supp_P}
$P$ uses $d=192$ and $N=6$ BiMamba blocks; each token holds the root offsets of two frames.
Its actor stream is the actor's 445 body and hand features, and its seed is the first reactor frame and root offset.
\subsection{Body module $B$}
\label{sec:supp_B}

\noindent\textbf{Architecture.}
$B$ uses $d=256$ and $N=8$ BiMamba blocks. Its actor stream has 265 channels per frame.
On Seamless Interaction, the 64-dimensional speech features of both participants are added (393 channels) and jointly dropped in 30\% of the training samples.
Before each block, $B$ also re-injects the actor tokens through a gated two-layer MLP, $\mathbf{h}_l\leftarrow\mathbf{h}_l
+a_l\,\mathrm{MLP}_l([\mathbf{h}_l\,\|\,\mathbf{K}])$, with a learned gate $a_l$.

\noindent\textbf{VAE objective.}
The body VAE, adapted from ReactionMamba~\cite{beg2026reactionmamba}, is trained before the flow model with
\begin{equation}
\mathcal{L}_{\mathrm{VAE}}=\mathcal{L}_{\mathrm{rec}}
+10^{-3}\mathcal{L}_{\mathrm{KL}}+0.5\,\mathcal{L}_{\mathrm{vel}}
+0.05\,\mathcal{L}_{\mathrm{skate}}.
\end{equation}
$\mathcal{L}_{\mathrm{rec}}$ is the $L_1$ reconstruction of positions,
rotations, velocities, and contacts.
$\mathcal{L}_{\mathrm{vel}}$ is the $L_1$ error of consecutive-frame
differences.
$\mathcal{L}_{\mathrm{KL}}$ is the KL divergence to a standard Gaussian.
$\mathcal{L}_{\mathrm{skate}}$, inspired by ARDY~\cite{zhao2026ardy},
is the mean reconstructed speed of the foot joints during GT contact.
The VAE is then frozen.

\noindent\textbf{Fixed placement during sampling.}
Each flow token of $B$ also holds the root offsets of its two frames (six channels).
During training, these channels hold the GT offsets, are never noised, and are excluded from the flow-matching loss.
At inference, the offsets predicted by $P$ (or the fixed placement on Seamless Interaction) are written into them before every Euler step and after the last one.

\noindent\textbf{Dual classifier-free guidance.}
On Inter-X and InterHuman, generation is guided separately by the actor tokens $\mathbf{K}$ and by the intent $\mathbf{e}$:
\begin{equation}
\begin{aligned}
\hat{\mathbf{v}}={}&\hat{\mathbf{v}}_{\varnothing,\varnothing}
+w_a\big(\hat{\mathbf{v}}_{\mathbf{K},\varnothing}
-\hat{\mathbf{v}}_{\varnothing,\varnothing}\big)\\
&+w_s\big(\hat{\mathbf{v}}_{\mathbf{K},\mathbf{e}}
-\hat{\mathbf{v}}_{\mathbf{K},\varnothing}\big),
\end{aligned}
\end{equation}
where $\varnothing$ denotes the learned null values; the seed and the offsets are kept in every evaluation.
We use $w_a=2.0$ on both datasets, and $w_s=2.5$ on Inter-X and $w_s=2.0$ on InterHuman.
On Seamless Interaction, $B$ is sampled without guidance.

\noindent\textbf{Conversational regularization.}
On Seamless Interaction, the model periodically generates a few training windows with the 25-step sampler during training.
Gradients flow through the last two steps and the frozen VAE decoder.
Two regularizers act on these generated windows.
The arm-range loss
$\mathcal{L}_{\mathrm{arm}}=\operatorname{mean}_j
\max(0,e_j-\hat e_j)/(e_j+10^{-3})$
compares the low-frequency ($0.05$--$1$\,Hz) amplitude $\hat e_j$ of the
generated shoulder and elbow rotations and wrist positions with that of
the VAE reconstruction of the GT window, $e_j$.
The listening-head loss $\mathcal{L}_{\mathrm{head}}
=\max(0,\bar v-v_0)/v_0$ penalizes the mean head-yaw speed $\bar v$
while the reactor is silent, above a threshold $v_0$: the 75th percentile of this statistic on GT training windows.
Table~\ref{tab:sl_BH_abl} reports their effect.

\subsection{Hand module $H$}
\label{sec:supp_H}

$H$ uses $d=320$ and $N=10$ BiMamba blocks; each token holds two frames
of 180 finger rotations.
Its actor stream contains the actor's body and hands.
Its reactor stream contains the GT body during training and the output
of $B$ at inference.
On Seamless Interaction, the speech features of each participant are
added to that participant's stream and jointly dropped in 30\% of the
training samples.
The seed $\mathbf{s}_H=[\mathbf{m}^{r}_1\,\|\,\mathbf{o}_1]$ holds the
first reactor frame, including its joint velocities, and the first 
offset.

\noindent\textbf{Hand--speech synchrony.}
Let $\hat{\mathbf{h}}_i$ be the generated finger rotations, taken from
$\hat{\mathbf{x}}_1$, and $\varepsilon_i$ the log-RMS energy of the
reactor's audio.
Over five-frame spans, we compute the finger motion
$\hat d_i=\|\hat{\mathbf{h}}_{i+5}-\hat{\mathbf{h}}_i\|_2$ and the mean
energy $\bar\varepsilon_i=\frac{1}{5}\sum_{k=1}^{5}\varepsilon_{i+k}$.
Their Pearson correlation over the window is
$\hat r_H=\operatorname{corr}_i(\hat d_i,\bar\varepsilon_i)$, and $r_H$
is the same correlation for the GT fingers.
The loss
\begin{equation}
\mathcal{L}^{H}_{\mathrm{sync}}=\max(0,\,r_H-\hat r_H)
\label{eq:h_sync}
\end{equation}
penalizes a correlation below the GT one.
It is applied only when the speech input is not dropped.

\subsection{Face module $F$}
\label{sec:supp_F}

$F$ uses $d=192$ and $N=6$ BiMamba blocks; each token holds two frames
of 53 facial parameters.
Its actor stream combines the actor's 445 body and hand features with
the actor's 53 facial parameters and their validity flag.
A second stream contains the reactor body.
The speech features of each participant are added to that participant's
stream and jointly dropped in 30\% of the training samples.
The seed holds the reactor's first body frame (GT in training, generated
at inference), first face, and root offset.
Its face part is set to zero when the first face is not tracked or the
reactor is speaking.
Training uses every frame with a tracked face, speaking frames included.

\noindent\textbf{Facial objectives.}
Three losses are computed from $\hat{\mathbf{x}}_1$, with
$\hat{\mathbf{f}}_i$ and $\mathbf{f}_i$ the predicted and GT facial
parameters.
The temporal loss supervises the change of the face between the two frames of each token and between frames two steps apart,
\begin{equation}
\mathcal{L}_{\Delta}=\sum_{\tau=1}^{2}
\operatorname{mean}_{i\in\mathcal{P}_\tau}
\big((\hat{\mathbf{f}}_{i+\tau}-\hat{\mathbf{f}}_i)
-(\mathbf{f}_{i+\tau}-\mathbf{f}_i)\big)^2,
\end{equation}
where $\mathcal{P}_1$ holds the first frame of each token and
$\mathcal{P}_2$ every frame, keeping only pairs with a tracked face at
both ends.
The jaw loss $\mathcal{L}_{\mathrm{jaw}}$ is the mean squared error of
the three jaw parameters on the speaking frames.
The synchrony loss compares the Pearson correlation, over the speaking
frames, between the jaw opening $\hat\phi_i$ and the speech energy
$\varepsilon_i$ with the same correlation $r_F$ for the GT face:
\begin{equation}
\mathcal{L}^{F}_{\mathrm{sync}}=\max(0,\,r_F-\hat r_F).
\label{eq:f_losses}
\end{equation}
The jaw and synchrony losses are applied only when the speech input is
not dropped.

\subsection{Training and inference}
\label{sec:train}

\noindent\textbf{Objectives.}
\begin{equation}
\begin{aligned}
\mathcal{L}_P &= \mathcal{L}_{\mathrm{FM}}+0.1\,\mathcal{L}_I,\\
\mathcal{L}_B &= \mathcal{L}_{\mathrm{FM}}+0.1\,\mathcal{L}_I
+\mathbf{1}_{\mathrm{SL}}\big(\mathcal{L}_{\mathrm{arm}}
+0.032\,\mathcal{L}_{\mathrm{head}}\big),\\
\mathcal{L}_H &= \mathcal{L}_{\mathrm{FM}}+0.1\,\mathcal{L}_I
+0.5\,\mathbf{1}_{\mathrm{SL}}\mathcal{L}^{H}_{\mathrm{sync}},\\
\mathcal{L}_F &= \mathcal{L}_{\mathrm{FM}}+0.1\,\mathcal{L}_I
+\mathcal{L}_{\Delta}+\mathcal{L}_{\mathrm{jaw}}
+0.5\,\mathcal{L}^{F}_{\mathrm{sync}},
\end{aligned}
\label{eq:total}
\end{equation}
where $\mathbf{1}_{\mathrm{SL}}$ is one on Seamless Interaction and zero otherwise; $F$ is used on Seamless Interaction only.

\noindent\textbf{Optimization.}
We use AdamW~\cite{loshchilov2019adamw} with weight decay $5\times10^{-4}$, gradient clipping at $1.0$, and dropout $0.1$.
The learning rate rises linearly to $2\times10^{-4}$ and then decays to 5\% of this peak.
On Seamless Interaction, $B$ trains its actor-reinjection layers at this rate and its other parameters at one tenth of this rate.
Table~\ref{tab:train_settings} gives the batch sizes and training lengths.
An exponential moving average of the parameters (decay $0.999$) is used at inference.

\noindent\textbf{Checkpoints.}
On Inter-X and InterHuman, checkpoints are chosen on the validation split ($P$ is the last checkpoint on InterHuman).
On Seamless Interaction, every module and every baseline uses its last checkpoint.

\begin{table}[t]
\centering
\scriptsize
\setlength{\tabcolsep}{4pt}
\caption{Batch size and training epochs per module. The body VAE is
trained before $B$.}
\label{tab:train_settings}
\begin{tabular}{@{}lccccc@{}}
\toprule
& VAE & $P$ & $B$ & $H$ & $F$ \\
\midrule
Inter-X & 250 & 150 & 400 & 400 & -- \\
InterHuman & 250 & 300 & 300 & -- & -- \\
Seamless Interaction & 300 & -- & 600 & 600 & 600 \\
\midrule
Batch size & 32 & 64 & 64 (16) & 64 (32) & 32 \\
\bottomrule
\end{tabular}

\smallskip
{\scriptsize Batch sizes in parentheses: Seamless Interaction.}
\end{table}

\noindent\textbf{Post-processing.}
On Inter-X and InterHuman, no filtering is applied.
On Seamless Interaction, the body generated by $B$ is smoothed with a nine-frame second-order Savitzky--Golay
filter~\cite{savitzky1964smoothing}, as Seamless Interaction does for its 6D rotations~\cite{agrawal2025seamless}.
Joint positions, velocities, and contacts are recomputed from the smoothed rotations, and this body conditions $H$ and $F$. The finger rotations of $H$ are smoothed with the same filter and then a 15-frame one; the face is not filtered.

\section{Datasets and splits}
\label{sec:supp_datasets}

\noindent\textbf{Inter-X}~\cite{xu2024interx} provides SMPL-X body and hand motion, interaction categories, captions, and actor--reactor annotations; we use 22 body and 30 finger joints. Actor and reactor follow the official preprocessing code.
We use the official split: 9,107 / 548 / 1,707 clips after length filtering.

\noindent\textbf{InterHuman}~\cite{liang2024intergen} provides 22-joint SMPL body motion~\cite{loper2015smpl} without hands, so $H$ is omitted; we use 5,609 / 546 / 1,087 clips.
Following Text2Interact~\cite{text2interact2026}, which classifies InterHuman captions into discrete interaction themes, we derive intent labels from the captions; instead of a language model, we use keyword rules (regular expressions) and no learned classifier.
The rules were written with the help of an AI coding assistant and refined by reading caption samples.
They map each of the three captions of a clip to one of 26 classes: 21 Inter-X classes and five InterHuman-specific ones (sparring, exercise, fencing, carrying, handing an object).
A clip keeps the majority class only when at least two captions agree. About half of the clips receive a label; the others are excluded from the intent loss.
A human blindly checked 25 random clips (20 labelled, 5 unlabelled): 24 decisions were judged correct and one uncertain. None was judged incorrect. The labels are used for training only.

\noindent\textbf{Seamless Interaction}~\cite{agrawal2025seamless} provides dyadic conversation videos, SMPL-H body and hand
parameters~\cite{romero2017mano}, separate audio tracks, and aligned transcripts.
We select paired windows of up to 16\,s with the UniLS-Talk index of UniLS~\cite{chu2025unils}, resample the motion to 25\,fps, and extract FLAME faces from the videos with TEASER~\cite{liu2025teaser}.

\noindent\textit{Roles.}
We keep the 37 of the 945 conversation prompts in the UniLS-Talk index that assign a support-seeking and a support-giving role.
The support giver is the reactor. The metadata does not say who played which role, so we read it from the
transcripts. The support giver mostly addresses the partner (``you'', ``your''), while the person seeking support mostly speaks about themselves (``I'', ``my'').
The participant with the larger ratio of second- to first-person pronouns is the reactor; recordings where the two ratios are too close are excluded. These decisions were then checked on the transcripts with the help of an AI assistant (Claude Opus),
which reversed 13 recordings and excluded 4.
A human then blindly checked the role assignments in 25 recordings, including 4 corrected and 2 excluded ones: For the 23 recordings retained in the final set, the reviewer confirmed 20 role assignments and found three unclear from the transcripts.

\noindent\textit{Split.}
The 1,697 paired windows (152 recordings, 63 pairs of people) are split by pairs of people, so that no pair or recording appears in two sets. Training has 1,273 windows (41 pairs), used in both directions;
validation has 144 windows (7 pairs); test has 280 windows (15 pairs). Four test windows have too few frames with a tracked face for the face metrics, so all Seamless results use the remaining 276 windows.
This is our own partition, not the dataset's official test set.

\noindent\textit{Placement.}
The two participants are recorded in separate spaces, so placement is not learned: $P$ is omitted, and both participants face each other at a fixed distance, for all compared methods.

\section{Conversational baselines}
\label{sec:supp_baselines}

\noindent\textbf{ConvoFusion.}
We retrain ConvoFusion~\cite{mughal2024convofusion} from scratch on our training windows with its published inputs: the reactor's own audio and transcript and the partner's transcript.
It trains for about 600 epochs, and we sample its last checkpoint with
its published settings.
The ``first pose'' row of Table~\mainTabSlBH{} of the main paper is the
same model given the reactor's first pose at test time, like our chain.

\begin{figure}[t]
\centering
\includegraphics[width=\columnwidth]{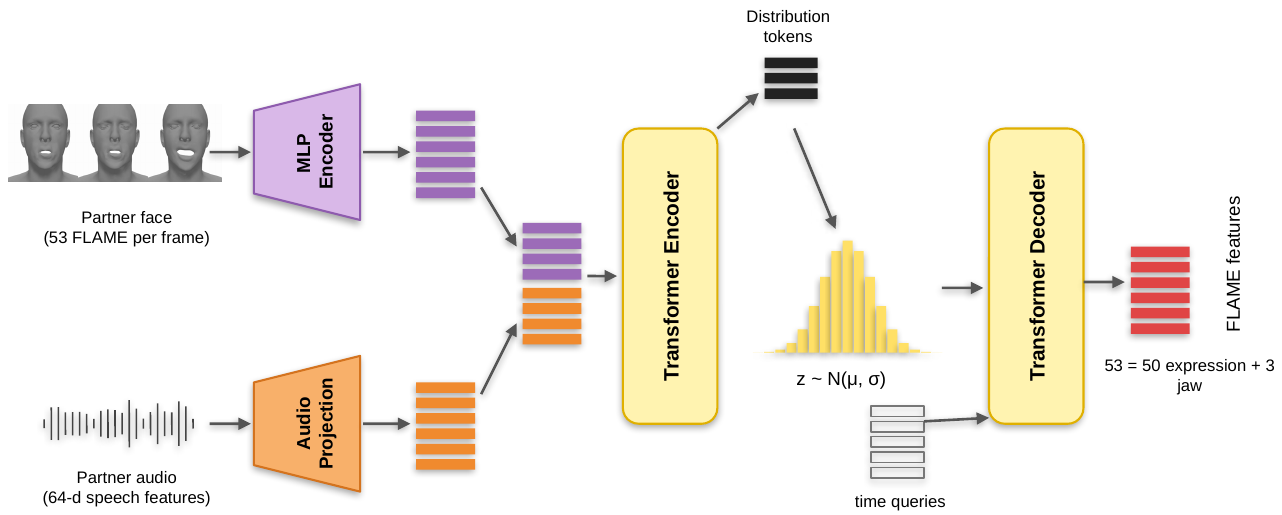}
\caption{Our adaptation of the REACT 2025 Trans-VAE
baseline~\cite{song2023react2023,song2025react}: only the inputs and the
output change. Loss $= 2\,\mathrm{MSE}(\text{expression})
+ 4\,\mathrm{MSE}(\text{jaw}) + 10^{-5}\,\mathrm{KL} + \text{diversity}$;
AdamW ($10^{-5}$), batch 4; 10 samples per window at test time.}
\label{fig:transvae}
\end{figure}

\noindent\textbf{Trans-VAE.}
Figure~\ref{fig:transvae} shows our adaptation of the REACT 2025
Trans-VAE baseline~\cite{song2023react2023,song2025react}.
The original model encodes the speaker's video with a 3-D convolutional
network.
Our data represents faces as FLAME parameters, so a per-frame MLP on the
partner's 53 FLAME parameters replaces this video encoder and produces
tokens of the same size.
The audio branch reads our 64-dimensional speech features, and the
output is the reactor's 53 FLAME parameters.
Everything else is the official code: the fusion, the VAE, the
transformer decoder, and the losses.
It is trained from scratch on our training windows for 100 epochs and
scored at its last checkpoint.

\section{Evaluation metrics}
\label{sec:metrics}

\noindent\textbf{Physical interactions.}
On Inter-X and InterHuman we use the released text--motion evaluators through the ARMFlow implementation~\cite{geng2026armflow}: the Inter-X evaluator and InterGen's InterCLIP evaluator~\cite{liang2024intergen},
with FID, R-precision, MM-Dist, and diversity as in HumanML3D~\cite{guo2022humanml3d}.
R-precision and MM-Dist use the reference captions, which our model does not read.
MPJPE (cm) and MPJVE (mm/frame) are the mean Euclidean errors of the reactor's joint positions and of their frame-to-frame differences against the GT reaction, on the subsets stated in the tables.

\noindent\textbf{Seamless body and hands.}
We train a GENEA EmbeddingNet autoencoder~\cite{kucherenko2023genea} with its published recipe on our training windows and freeze it. Its encoder maps a 90-frame chunk of standardized joint positions to a 32-dimensional code.
Generated and GT windows are cut into chunks with stride 45. With $(\boldsymbol{\mu},\boldsymbol{\Sigma})$ and
$(\boldsymbol{\mu}_{\mathrm{GT}},\boldsymbol{\Sigma}_{\mathrm{GT}})$ the
mean and covariance of the two code sets, 
\begin{equation}
\mathrm{FGD}=\|\boldsymbol{\mu}-\boldsymbol{\mu}_{\mathrm{GT}}\|_2^2
+\operatorname{tr}\!\big(\boldsymbol{\Sigma}
+\boldsymbol{\Sigma}_{\mathrm{GT}}
-2(\boldsymbol{\Sigma}\boldsymbol{\Sigma}_{\mathrm{GT}})^{1/2}\big).
\end{equation}
The inputs are the 22 body joints relative to the pelvis (body22) or the 30 finger joints relative to their wrist (hands30), with one autoencoder per input.
The upper FGD is the body22 FGD after the generated hips, spine and legs are replaced by the GT ones, so it scores the neck, head, and arms.

For the ablations, let $\theta_i$ be the head yaw relative to the chest and $v_i$ the head speed, in degrees per second.
A listening head shake is a turn of $\theta$ of at least $3^\circ$ that is reversed within 1\,s while the reactor is silent; we count them per minute of silence.
The head--speech correlation is the Pearson correlation between $v_i$ and the log-RMS energy of the reactor's audio, averaged over windows.
The finger speak/listen ratio divides the mean finger rotation speed while the reactor speaks by that while it is silent.
These values use our windows and autoencoders and are not comparable to published GENEA numbers.

\noindent\textbf{Seamless face.}
The REACT 2025 metrics~\cite{song2025react} are computed on the 53 FLAME parameters of the $S=10$ generated faces $\hat{\mathbf{f}}^{(s)}$ of each window.
Each window has an appropriate set $\mathcal{A}$: the GT faces of the windows that share its conversation prompt.
With CCC the concordance correlation coefficient, averaged over the 53 parameters, and DTW the dynamic-time-warping distance,
\begin{equation}
\begin{aligned}
\mathrm{FRC}&=\frac{1}{S}\sum_{s}\max_{\mathbf{f}\in\mathcal{A}}
\mathrm{CCC}(\mathbf{f},\hat{\mathbf{f}}^{(s)}),\\
\mathrm{FRD}&=\sum_{s}\min_{\mathbf{f}\in\mathcal{A}}
\mathrm{DTW}(\hat{\mathbf{f}}^{(s)},\mathbf{f}),\\
\mathrm{FRVar}&=\frac{1}{S}\sum_{s}\operatorname{mean}_{d}
\operatorname{Var}_{i}\,\hat f^{(s)}_{i,d},\\
\text{S-MSE}&=\frac{1}{S(S-1)}\sum_{s\neq s'}\operatorname{mean}_{i,d}
\big(\hat f^{(s)}_{i,d}-\hat f^{(s')}_{i,d}\big)^2,
\end{aligned}
\end{equation}
all averaged over the test windows.
FRDvs is the same mean squared difference between the seed-averaged faces of two different windows.
TLCC is the absolute lag, within $\pm2$\,s, that maximizes the cross-correlation between the generated face and the actor's face. FRD also rewards faces that barely move, so we read it together with
FRC. These values are not comparable to published REACT numbers.

\noindent\textbf{Intervals.}
The $\pm$ values of the Seamless tables are 95\% intervals over the generation runs on the fixed test set.

\section{Seamless Interaction ablations}
\label{sec:supp_sl_ablations}
\input{tabels/table5_seamless_body_hands_ablation}
\input{tabels/table7_seamless_face_ablation}
The Inter-X ablations are reported in Table~\mainTabInterXAbl{} of the main paper.
Table~\ref{tab:sl_BH_abl} changes one element of the body and hand chain at a time.
The arm-range and listening-head losses improve the upper FGD (0.207 against 0.260) and the head--speech correlation (0.121 against 0.087; GT 0.194), and lower the listening head shakes from 55.1 to 32.3 per minute (GT 10.9).
Without its audio, the head of the body module no longer follows the reactor's speech (correlation 0.005) and shakes more while the reactor listens (41.1 per minute).
Without its audio, the hand module loses about half of its speak/listen contrast (ratio 1.16 against 1.29; GT 1.37).
Table~\ref{tab:sl_face_abl} ablates the face module on the GT body: without the body input, FRC drops from 0.296 to 0.153 while the faces vary more between seeds (S-MSE 1.320 against 0.643); without audio on silent frames, FRC falls to 0.281.

\section{User study}
\label{sec:supp_user_study}

\begin{figure}[t]
\centering
\includegraphics[width=0.75\columnwidth]{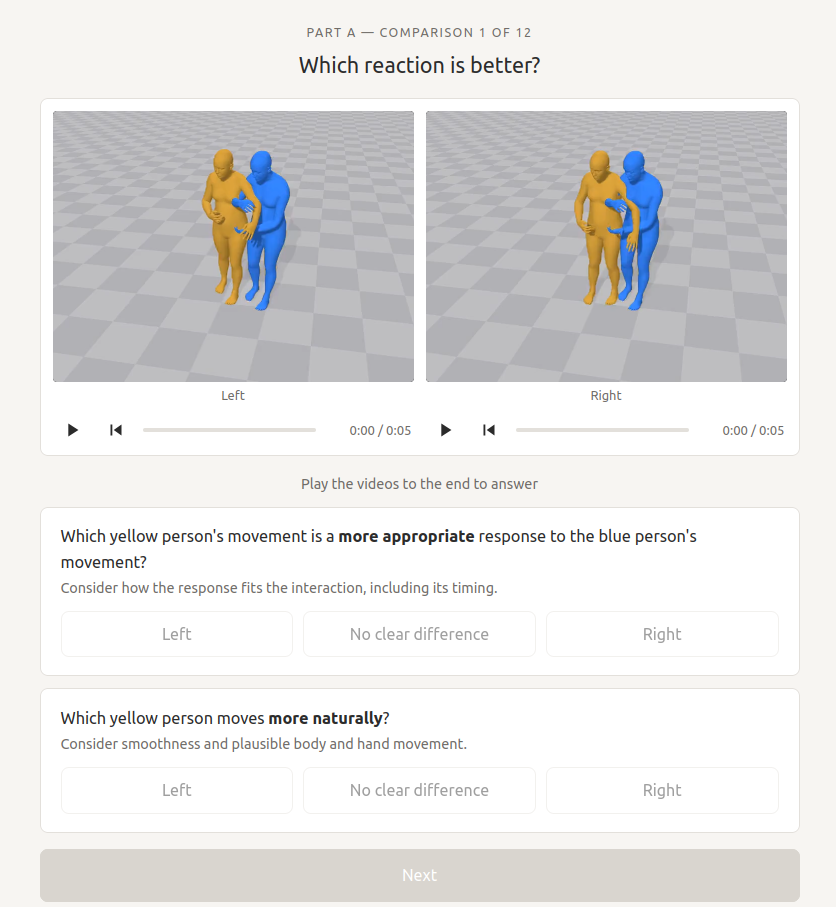}
\caption{Study page for Inter-X and InterHuman, as the raters saw it: the two reactions side by side without sound.}
\label{fig:study_partA}
\end{figure}

\begin{figure}[t]
\centering
\includegraphics[width=0.48\columnwidth]{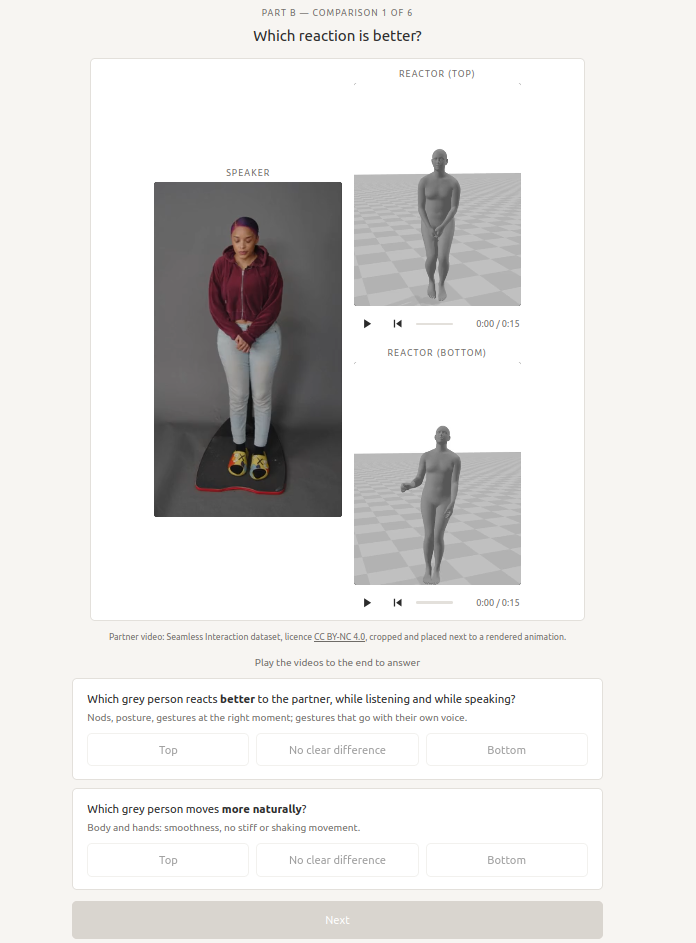}\hfill
\includegraphics[width=0.48\columnwidth]{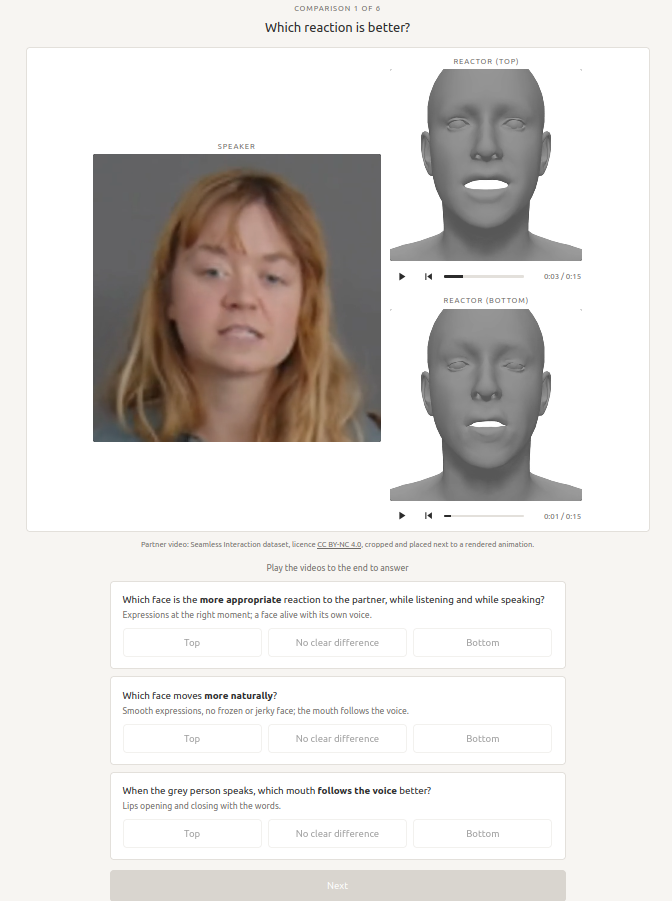}
\caption{Study pages for Seamless Interaction, as the raters saw them.
Left: main study, the partner's video with both voices and two reactor
bodies stacked. Right: face study, the partner's recorded face and two
reactor faces stacked. The partner video is from Seamless Interaction
(CC BY-NC 4.0), with the credit line shown on the page.}
\label{fig:study_partB}
\end{figure}

\noindent\textbf{Main study.}
Section~\mainSecUserStudy{} of the main paper reports the main study. It is a blinded A/B comparison with a ``no clear difference'' option, where each screen asks which reaction is more appropriate and which motion is more natural.
Six test clips per dataset were chosen on GT previews before any method was rendered.
All videos use one renderer, one camera per clip, and a neutral body shape for every row, the GT included; on Seamless Interaction, the partner's recorded video is shown with both voices next to the reactor as a grey mesh.
Figures~\ref{fig:study_partA} and~\ref{fig:study_partB} show the study pages.

\noindent\textbf{Face study.}
\begin{table}[t]
\centering
\caption{Face study (14 raters): share of judgements preferring our face
(Win), no difference (Tie), or the other face (Loss). Bold: the 95\%
confidence interval of the preference for our face excludes 50\%.}
\label{tab:user_study_face}
\resizebox{\columnwidth}{!}{%
\begin{tabular}{@{}llrrrrrrrrr@{}}
\toprule
& & \multicolumn{3}{c}{Appropriateness}
& \multicolumn{3}{c}{Naturalness}
& \multicolumn{3}{c}{Lip sync} \\
\cmidrule(lr){3-5} \cmidrule(lr){6-8} \cmidrule(lr){9-11}
Dataset & Ours vs. & Win & Tie & Loss & Win & Tie & Loss
& Win & Tie & Loss \\
\midrule
Seamless, face & Trans-VAE & \textbf{86\%} & 5\% & 10\% & \textbf{83\%} & 5\% & 12\%
& \textbf{88\%} & 2\% & 10\% \\
& GT & 36\% & 5\% & 60\% & 48\% & 5\% & 48\%
& 48\% & 14\% & 38\% \\
\bottomrule
\end{tabular}%
}
\end{table}
 A separate study compares the faces of the complete chain.
Each screen shows the partner's recorded face with both voices and two
reactor faces rendered as grey masks, frontal with a fixed head pose, so
that only the expression differs (Figure~\ref{fig:study_partB}, right).
Our face is generated by the full chain, from the body generated by $B$.
It is compared with Trans-VAE and with the GT face on the six Seamless
windows of the main study.
Three questions are asked per screen: which face is the more appropriate
reaction, which face moves more naturally, and, when the reactor speaks,
which mouth follows the voice better.
Table~\ref{tab:user_study_face} reports the results of 14 raters.
Our face is preferred over Trans-VAE on all three questions.
Against the GT face, it is not distinguished on naturalness and lip
synchrony, and is chosen less often as the more appropriate reaction.


\section{Qualitative results}
\label{sec:supp_qual}

\begin{figure*}[p]
\centering
\includegraphics[width=0.95\textwidth]{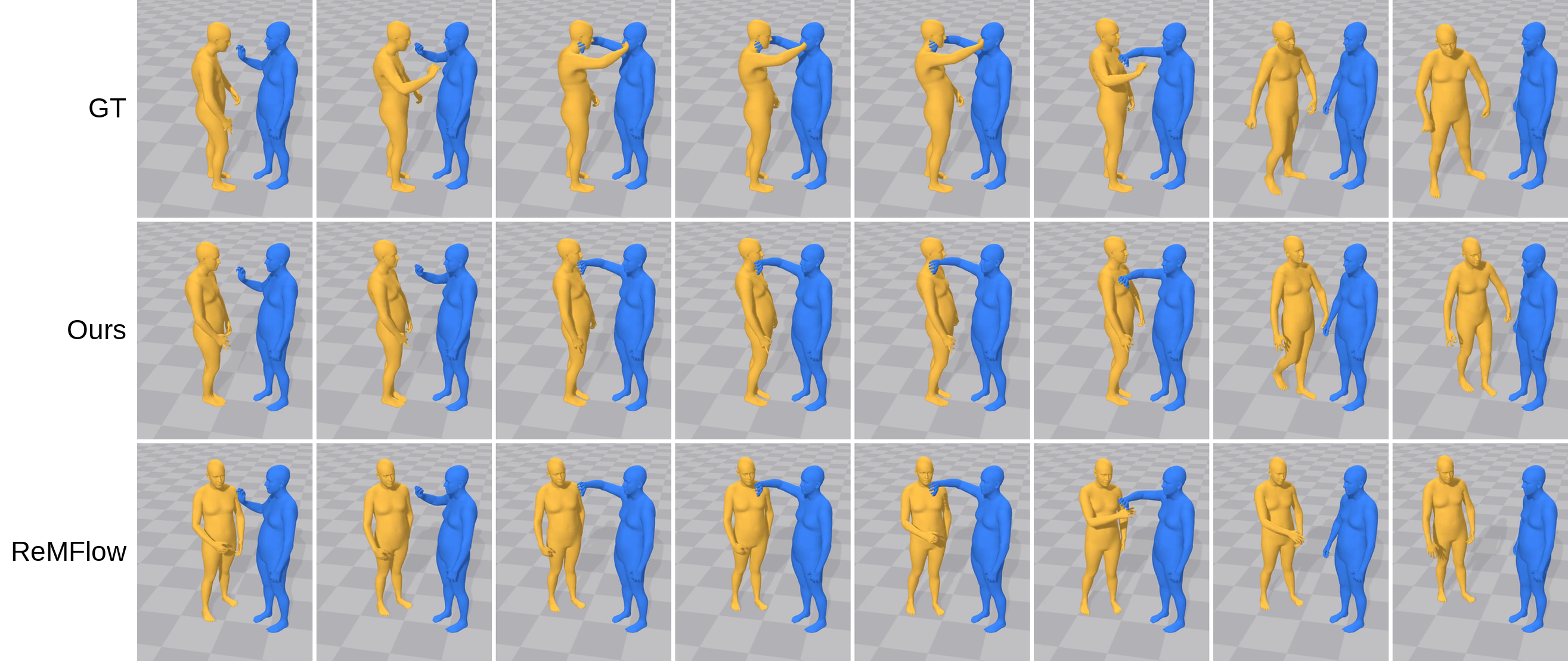}\\[10pt]
\includegraphics[width=0.95\textwidth]{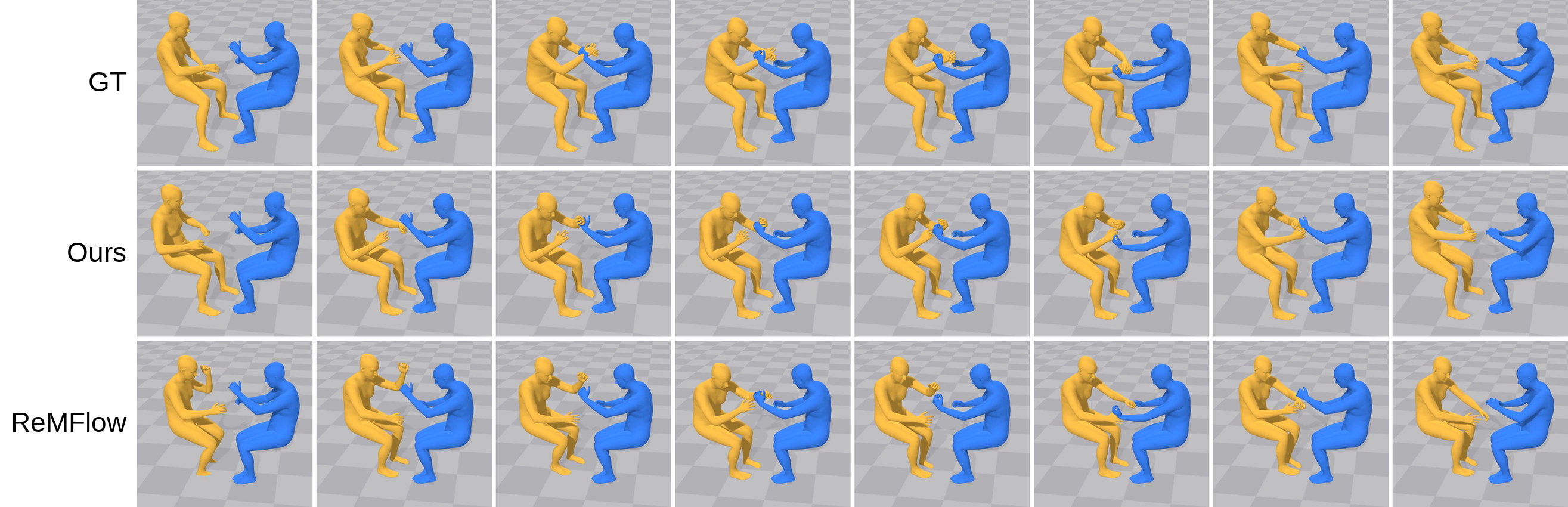}\\[10pt]
\includegraphics[width=0.95\textwidth]{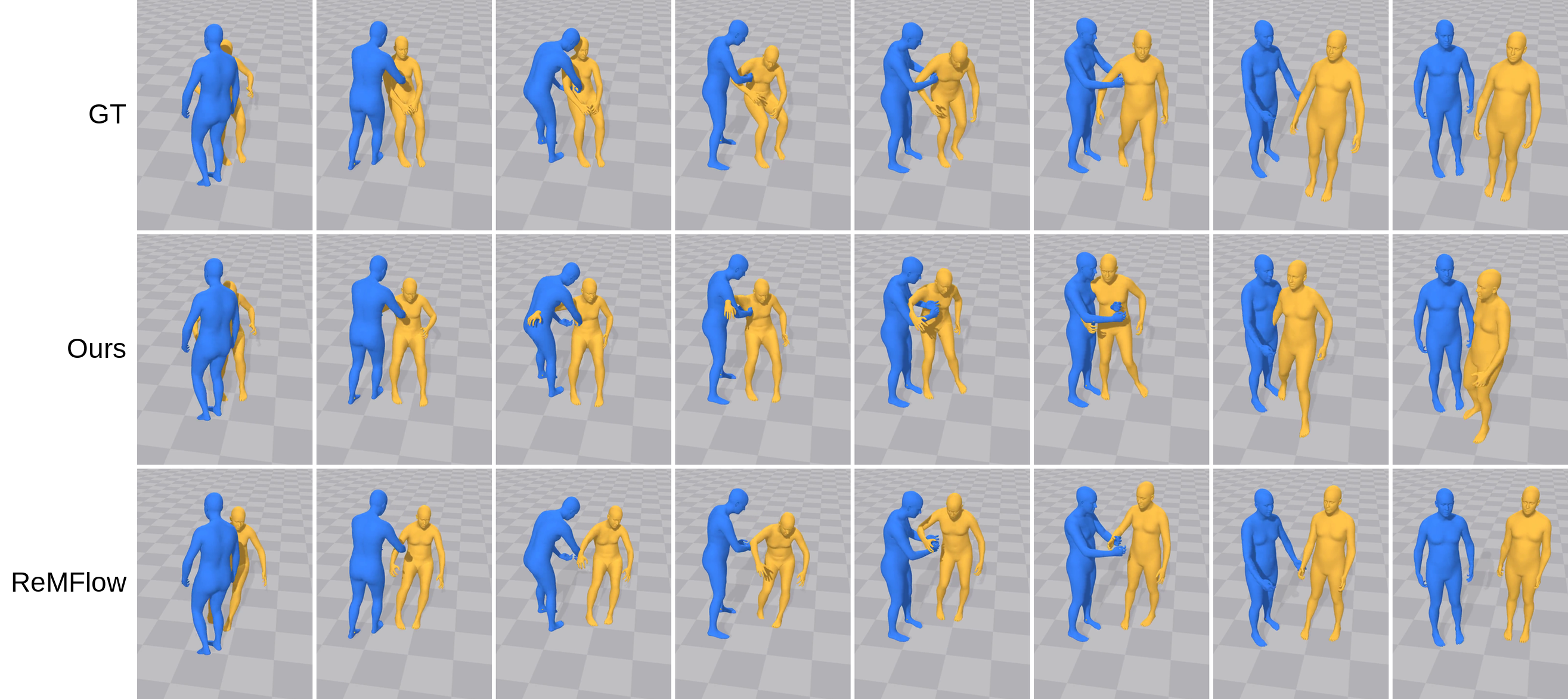}
\caption{Inter-X study clips: cover mouth, hand wrestling, help up (top to
bottom). Rows: GT reactor, ours, ReMFlow~\cite{geng2026armflow}; actor in
blue, reactor in yellow.}
\label{fig:qual_interx_a}
\end{figure*}

\begin{figure*}[p]
\centering
\includegraphics[width=0.95\textwidth]{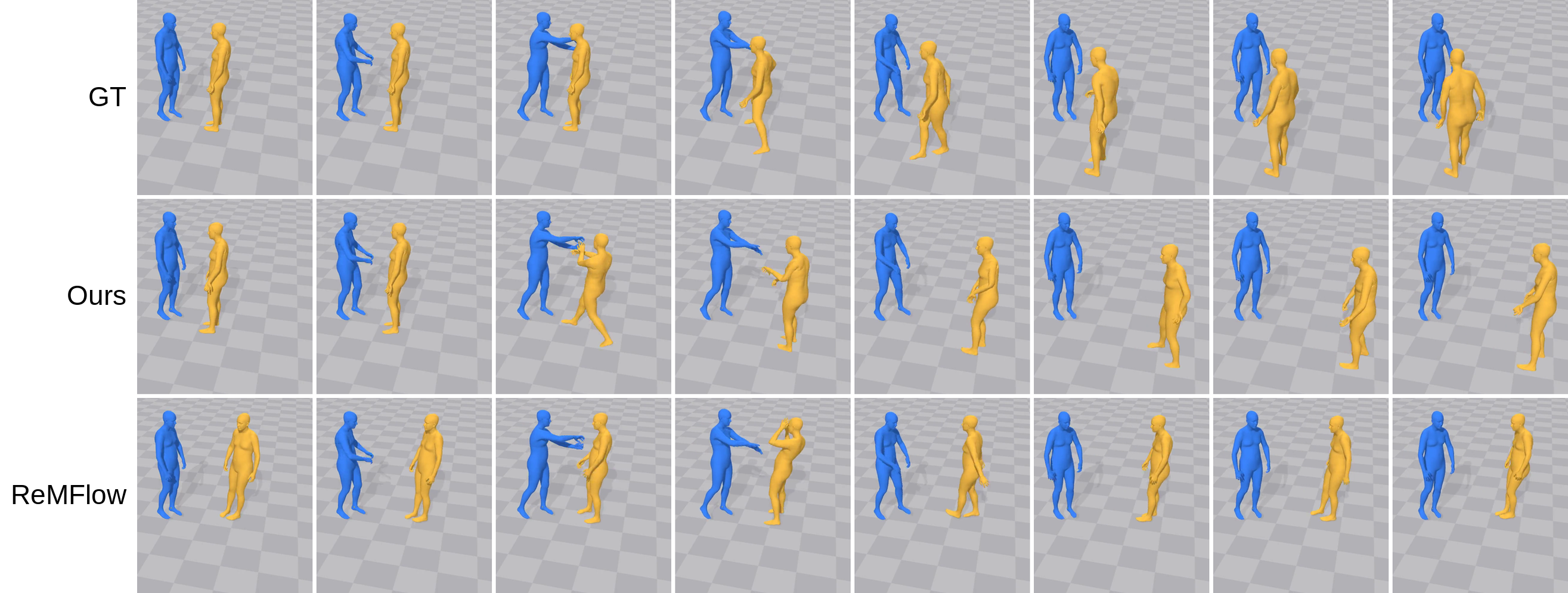}\\[10pt]
\includegraphics[width=0.95\textwidth]{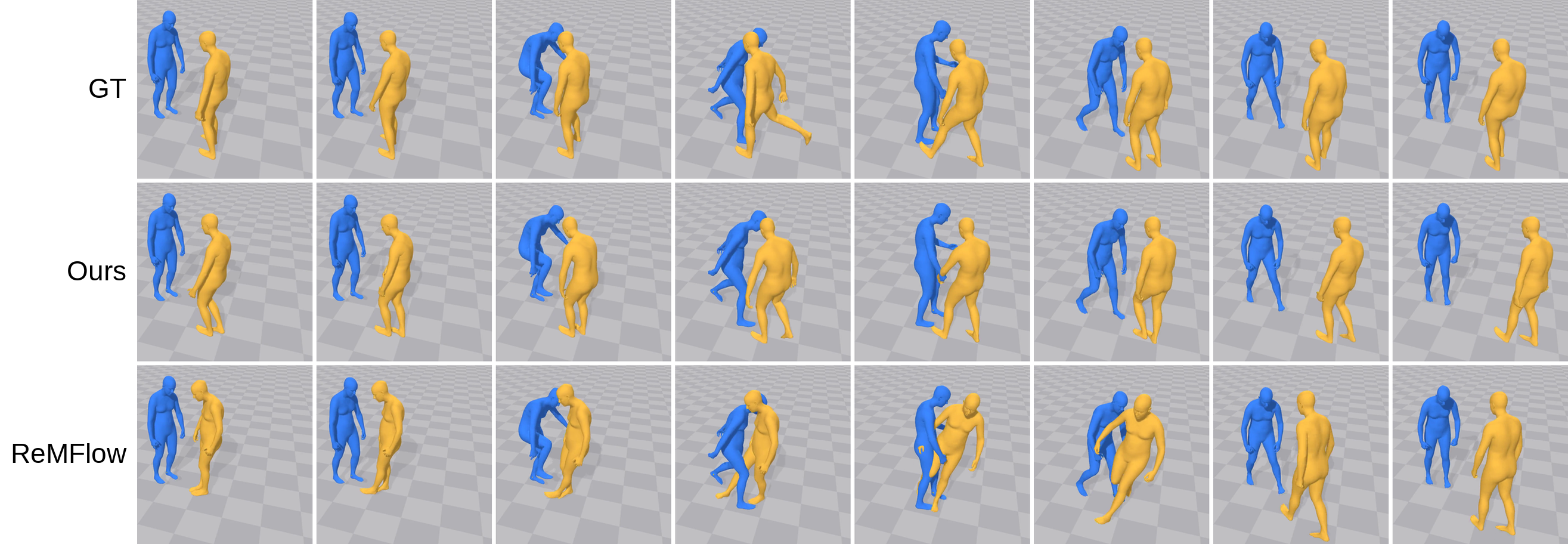}\\[10pt]
\includegraphics[width=0.95\textwidth]{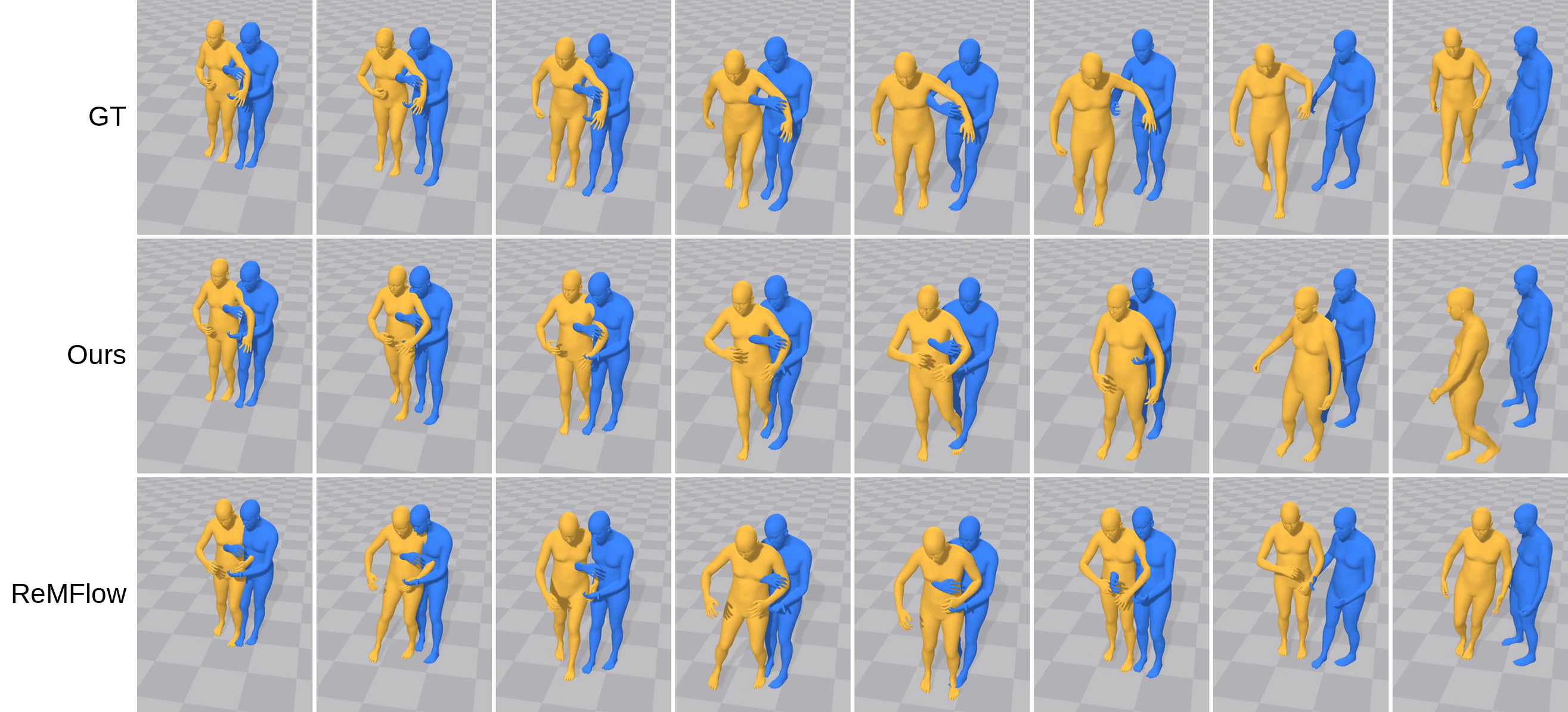}
\caption{Inter-X study clips: hit, step on foot, support with hand (top to
bottom). Same layout as Figure~\ref{fig:qual_interx_a}.}
\label{fig:qual_interx_b}
\end{figure*}

\begin{figure*}[p]
\centering
\includegraphics[width=0.95\textwidth]{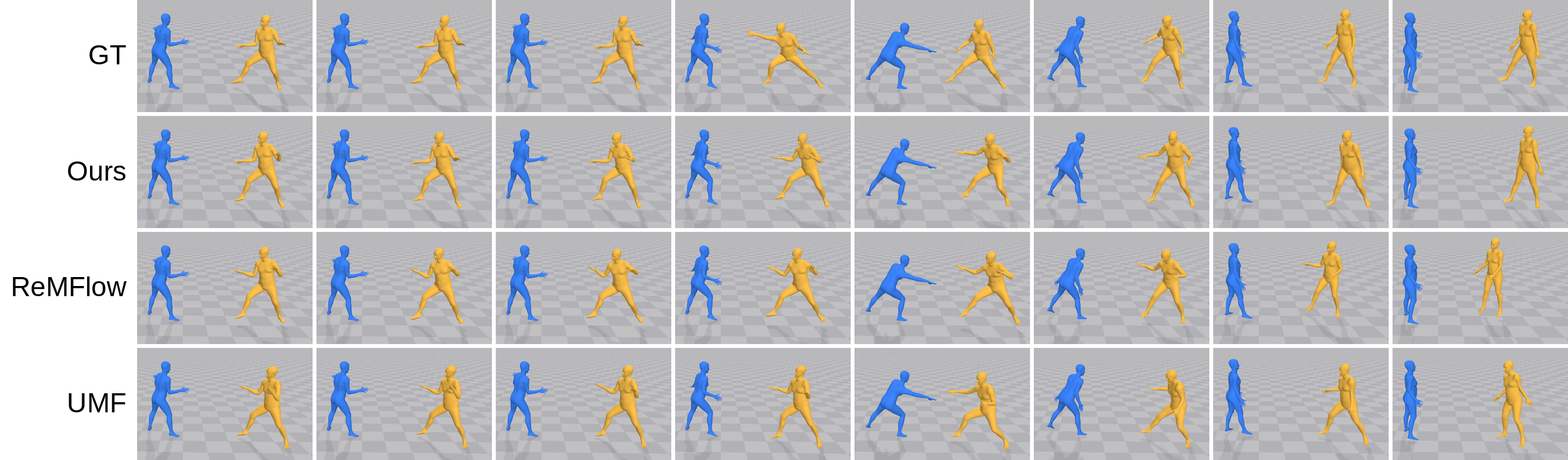}\\[10pt]
\includegraphics[width=0.95\textwidth]{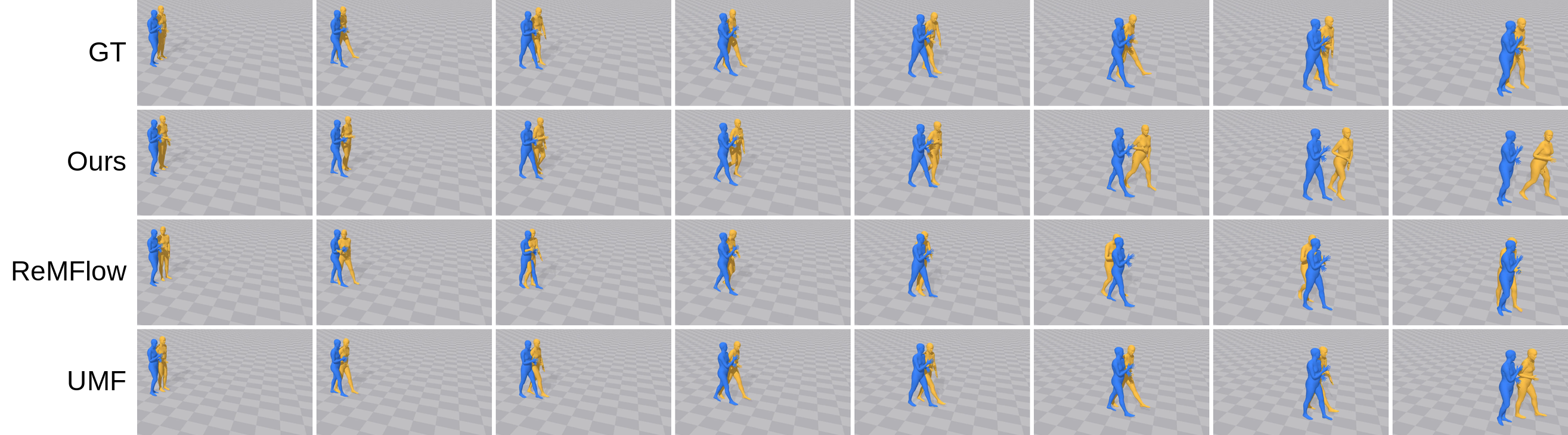}\\[10pt]
\includegraphics[width=0.95\textwidth]{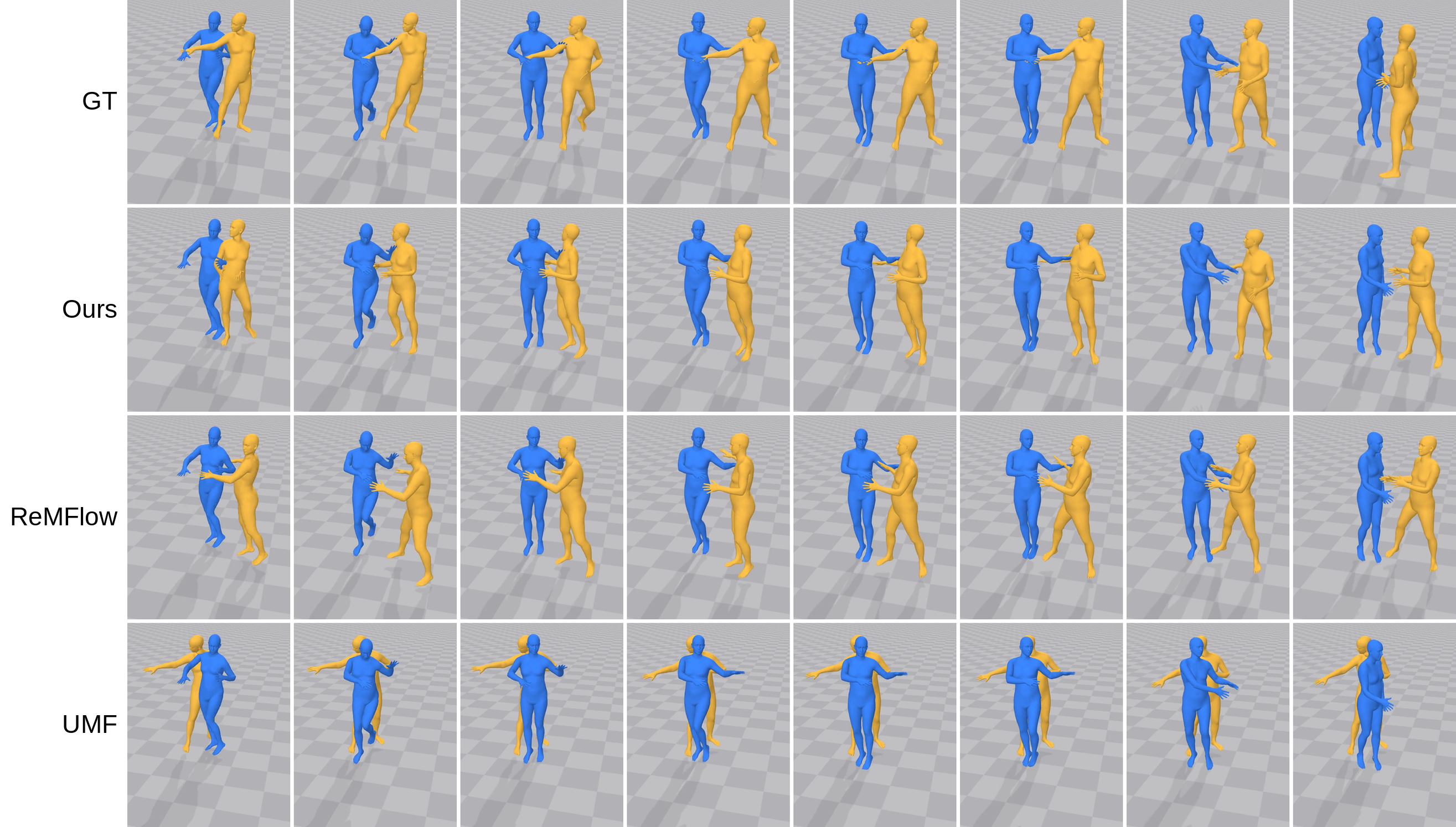}
\caption{InterHuman study clips: sequences 967, 1064, and 2209 (top to
bottom). Rows: GT reactor, ours, ReMFlow~\cite{geng2026armflow},
UMF~\cite{huang2026umf}; InterHuman has no finger data.}
\label{fig:qual_interhuman_a}
\end{figure*}

\begin{figure*}[p]
\centering
\includegraphics[width=0.95\textwidth]{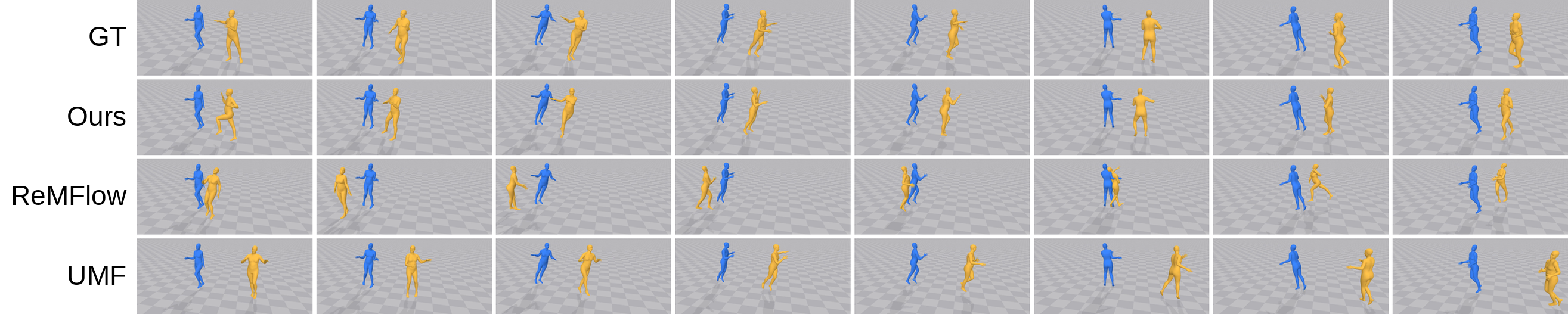}\\[10pt]
\includegraphics[width=0.95\textwidth]{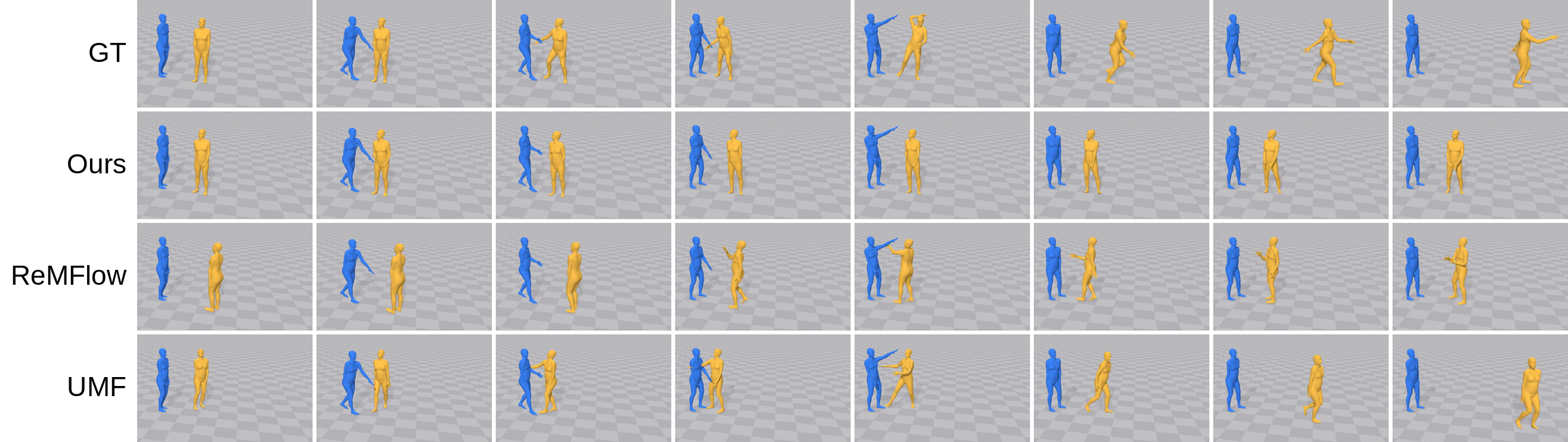}\\[10pt]
\includegraphics[width=0.95\textwidth]{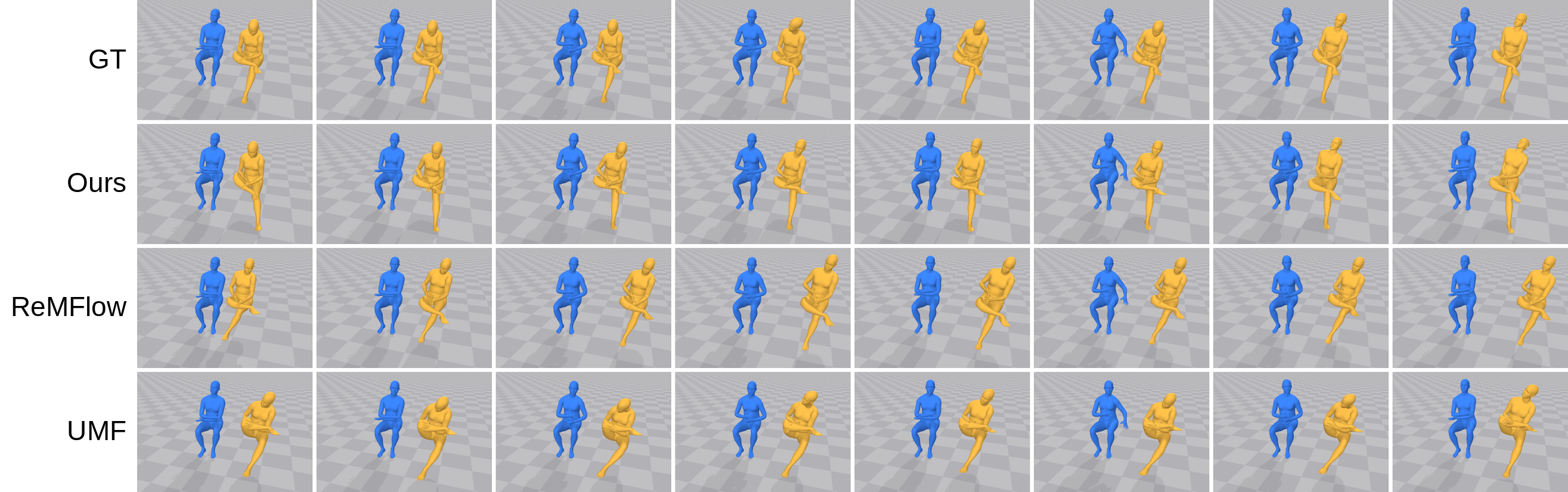}
\caption{InterHuman study clips: sequences 2260, 4038, and 4614 (top to
bottom). Same layout as Figure~\ref{fig:qual_interhuman_a}.}
\label{fig:qual_interhuman_b}
\end{figure*}

\begin{figure*}[p]
\centering
\includegraphics[width=0.48\textwidth]{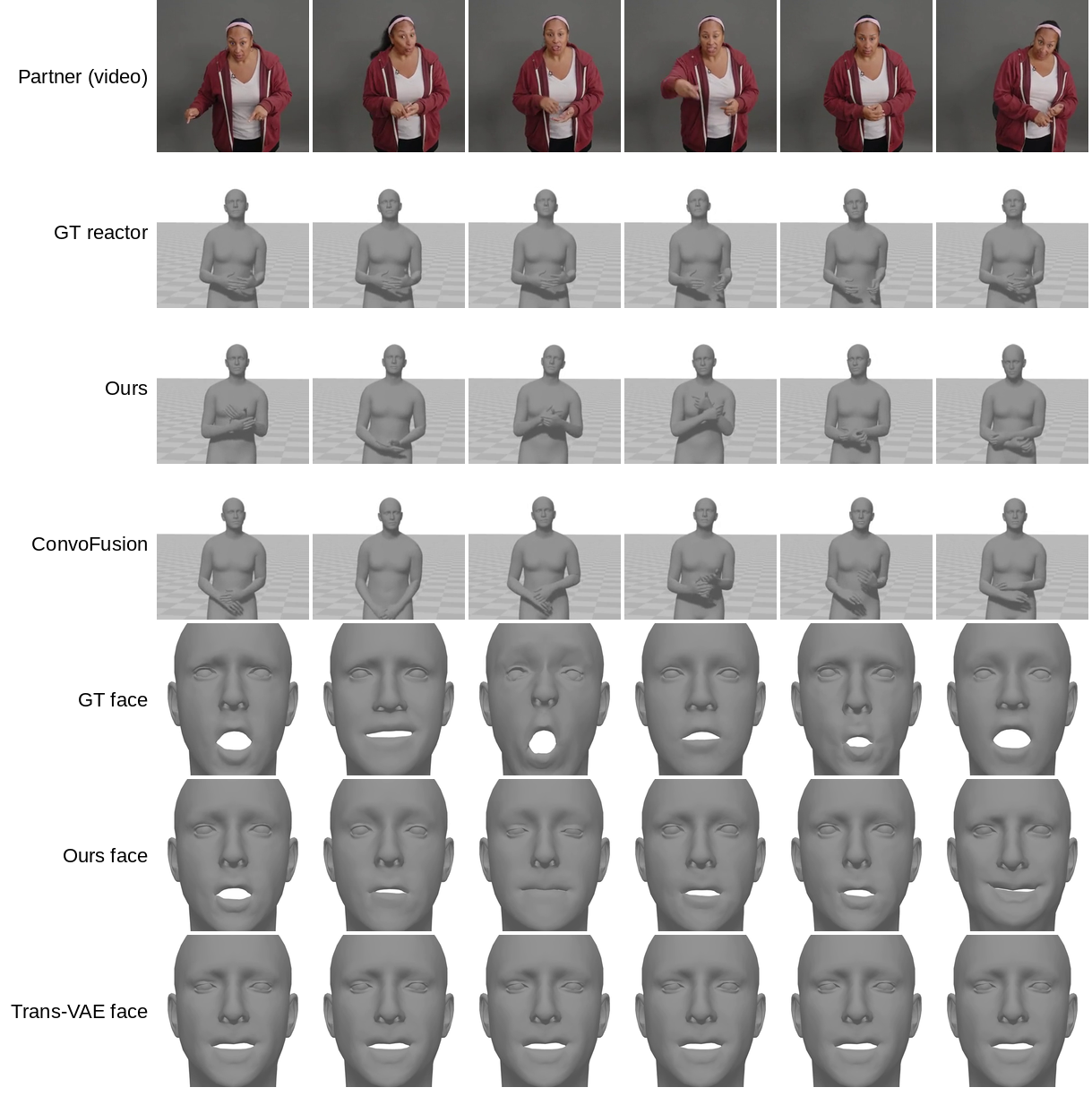}\hfill
\includegraphics[width=0.48\textwidth]{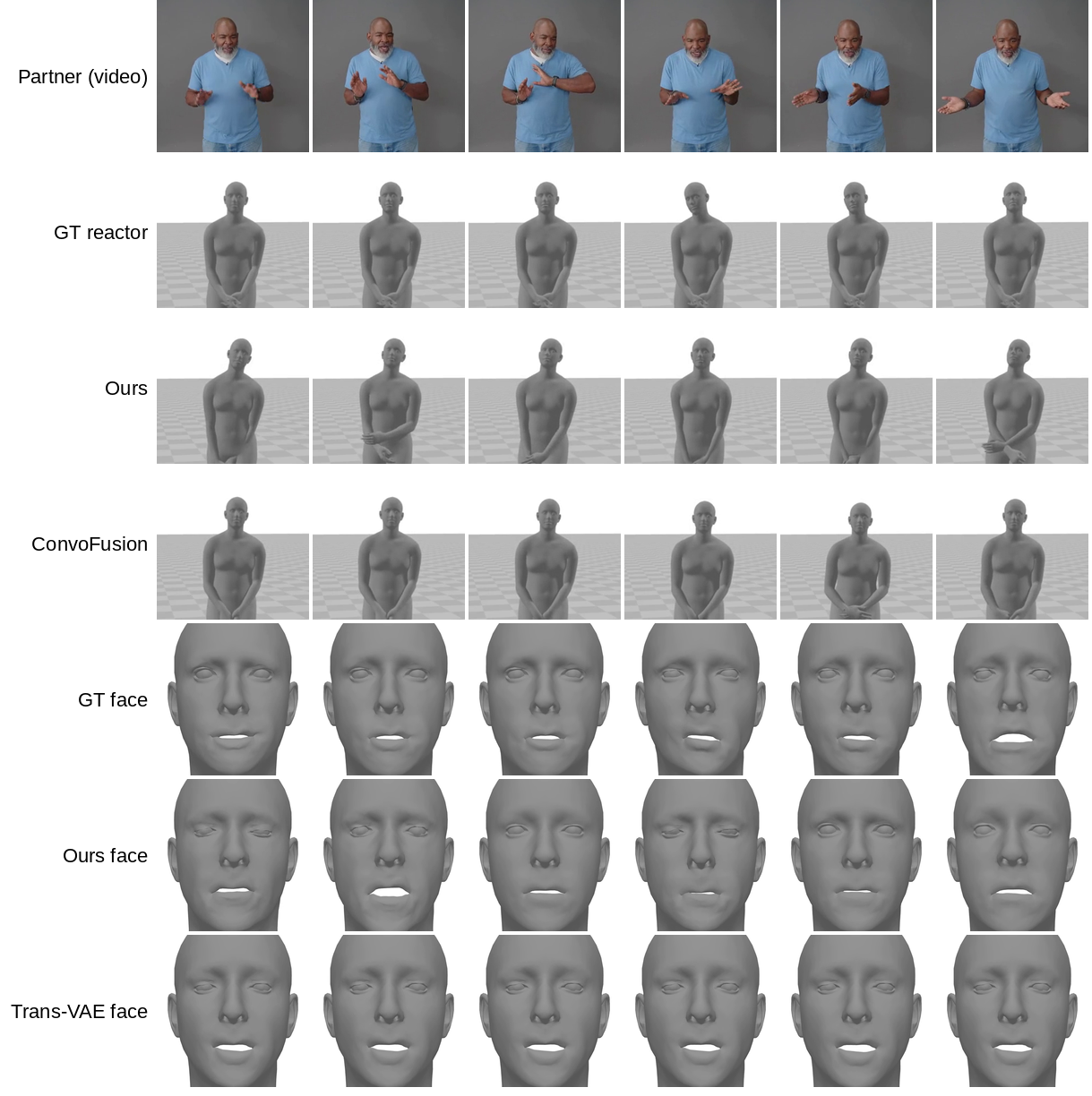}\\[10pt]
\includegraphics[width=0.48\textwidth]{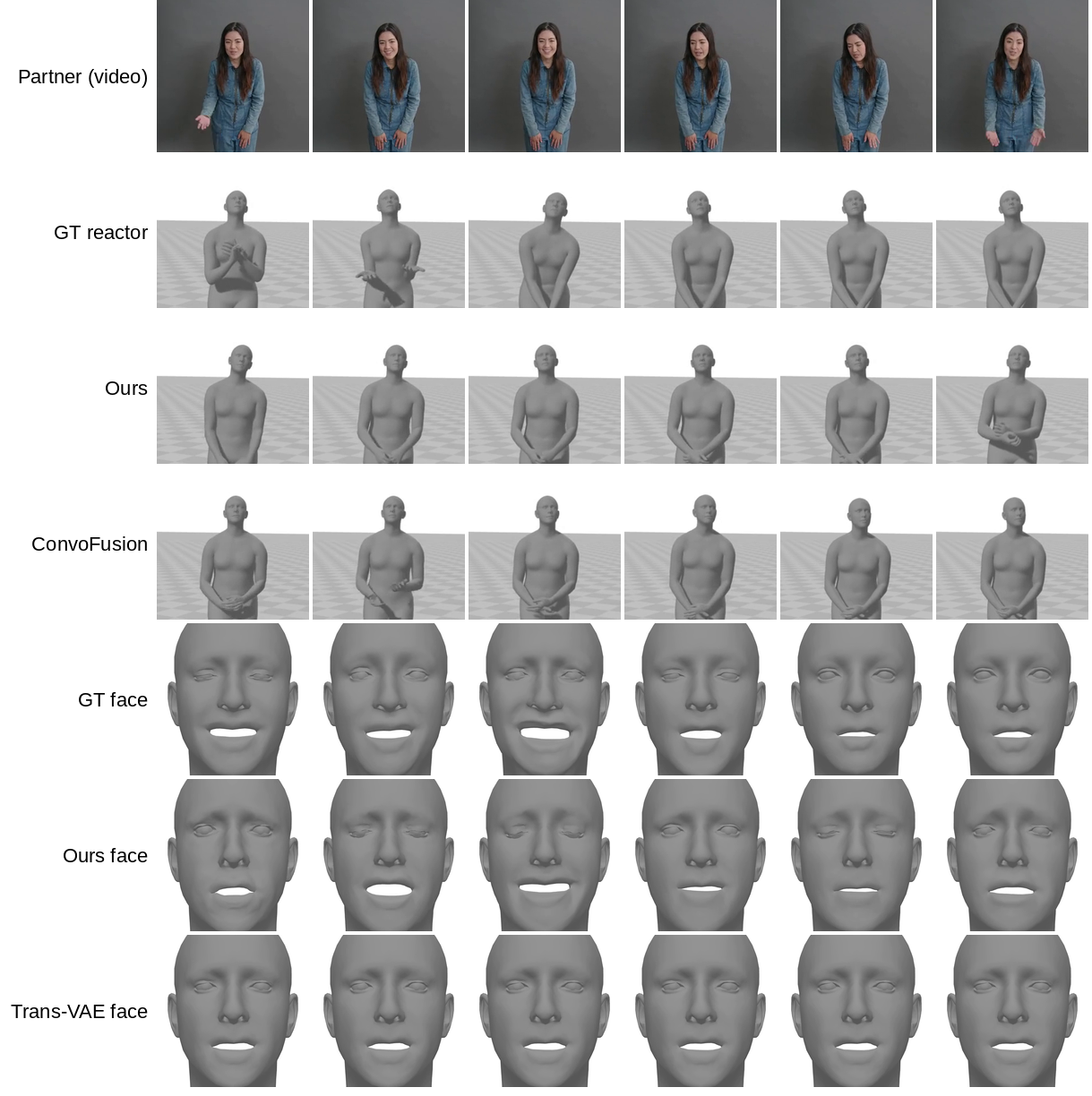}\hfill
\includegraphics[width=0.48\textwidth]{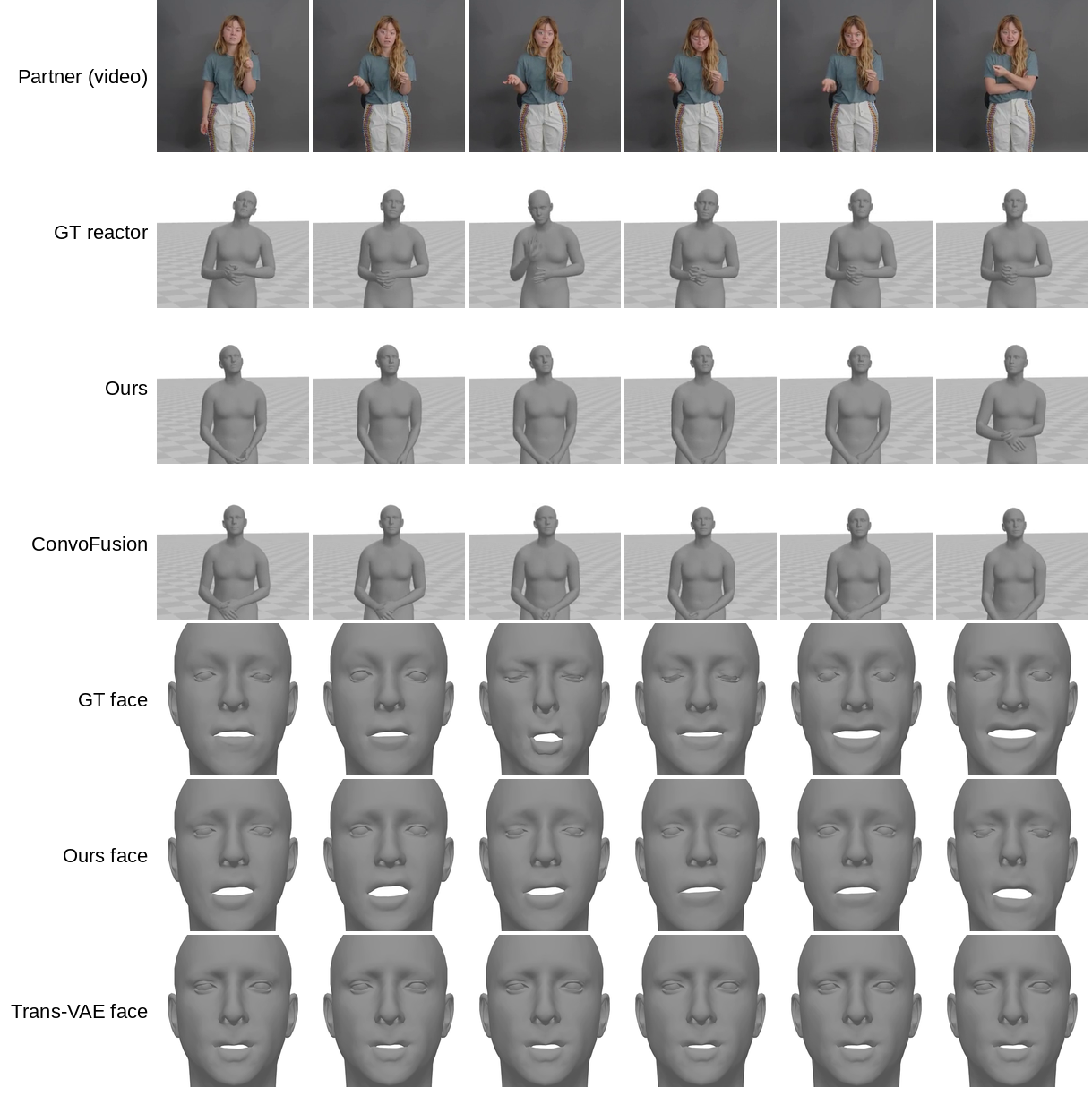}
\caption{Seamless Interaction study windows 1--4 (left to right, top to
bottom). Rows: the partner's recorded video; the reactor's upper body for
the GT, ours, and ConvoFusion~\cite{mughal2024convofusion} given the
first pose (display-only SMPL-X fit); then the face masks of the face
study for the GT, ours (full chain), and Trans-VAE~\cite{song2025react}.}
\label{fig:qual_seamless_a}
\end{figure*}

\begin{figure*}[p]
\centering
\includegraphics[width=0.48\textwidth]{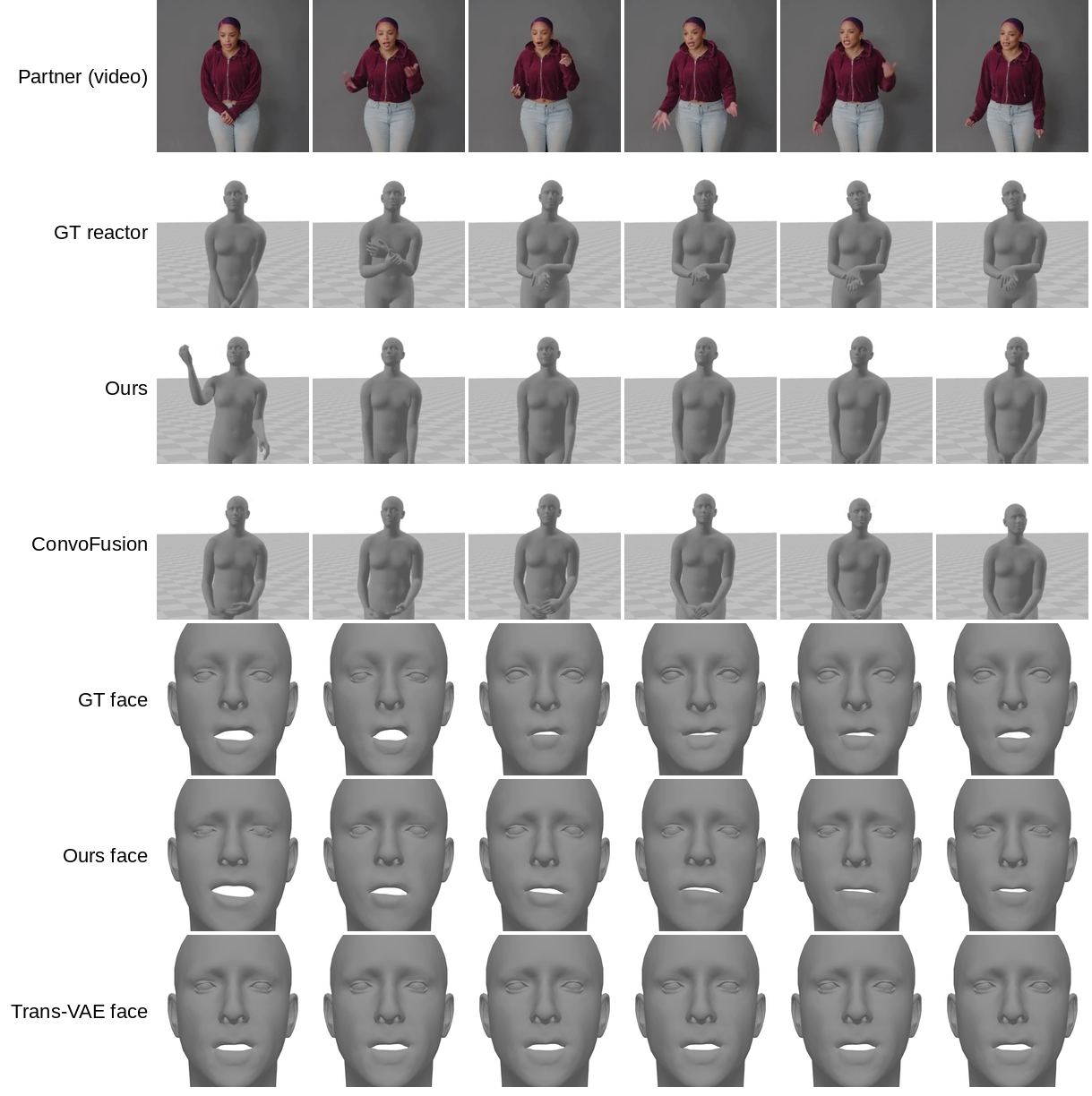}\hfill
\includegraphics[width=0.48\textwidth]{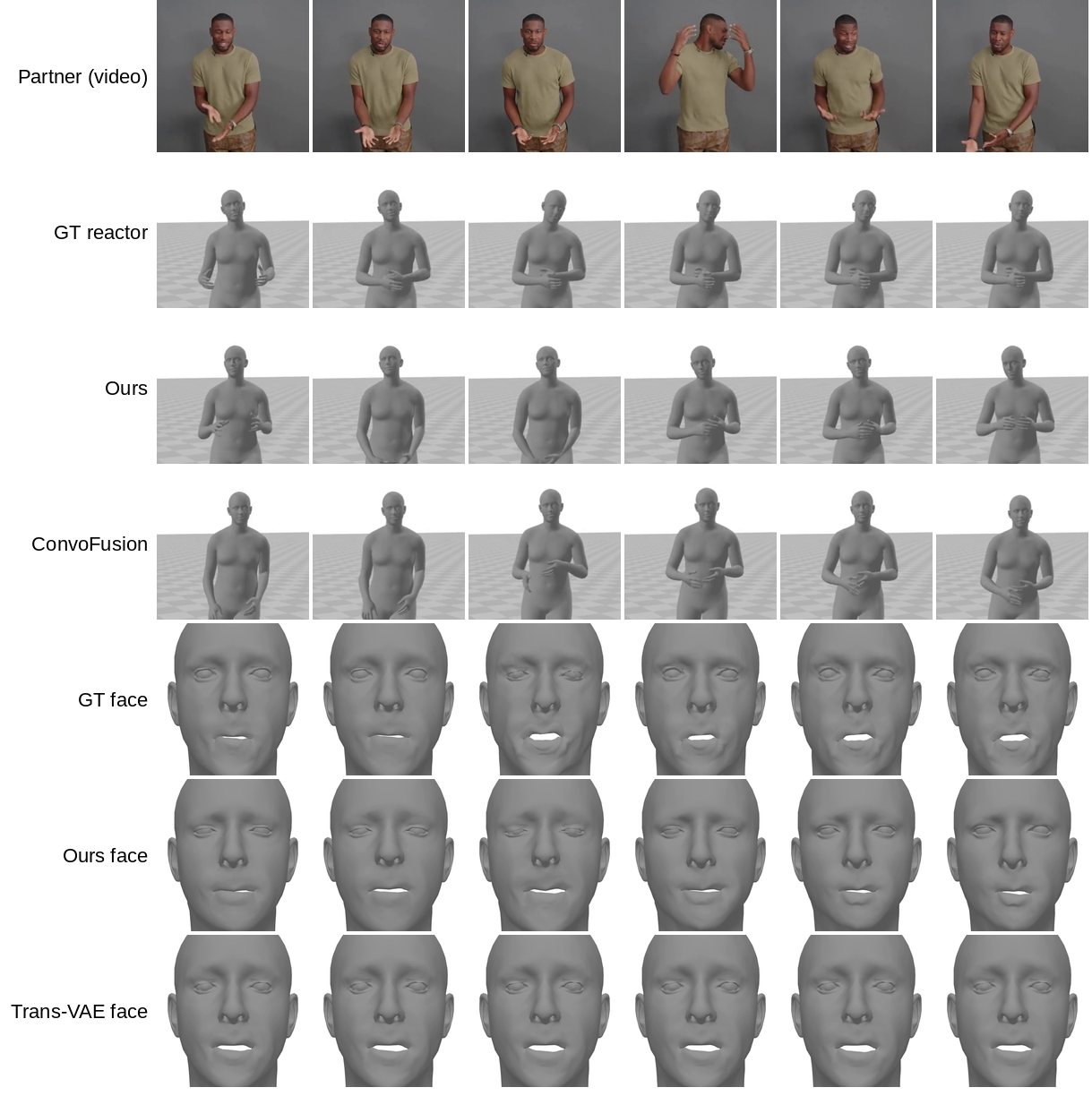}\\[10pt]
\includegraphics[width=0.48\textwidth]{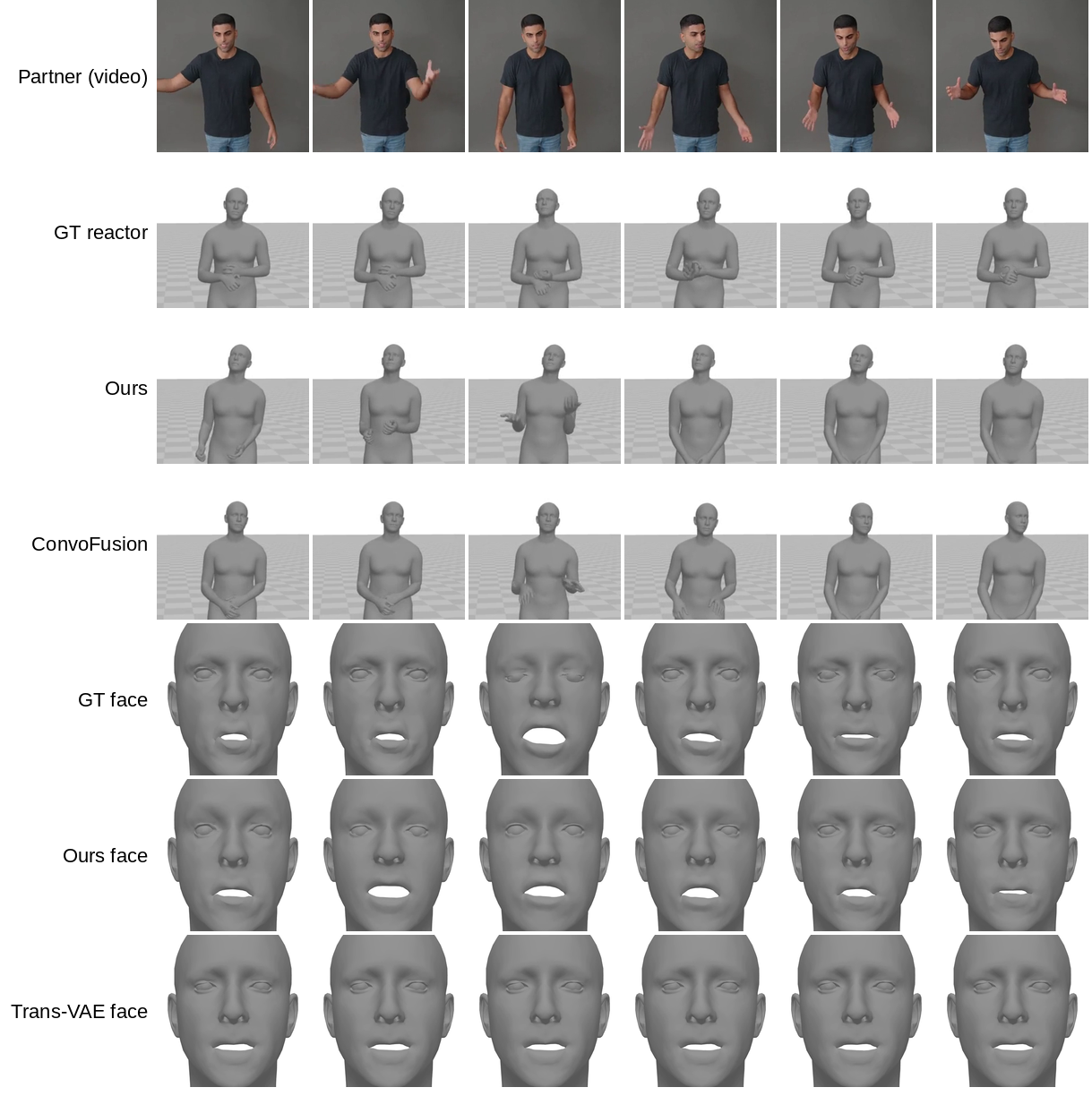}
\caption{Seamless Interaction study windows 5 and 6 (top) and window 7
(bottom), which was not shown to the raters. Same layout as
Figure~\ref{fig:qual_seamless_a}.}
\label{fig:qual_seamless_b}
\end{figure*}

Figures~\ref{fig:qual_interx_a}--\ref{fig:qual_seamless_b} show the clips of
the user study with frames taken from the supplementary videos; InterHuman sequence 2260 is the clip shown in Figure ~\mainFigQual{} of the main paper.
In every figure, the frames are evenly spaced and the same in every row.

 \clearpage

%% file: tabels/table5_seamless_body_hands_ablation.tex
\begin{table*}[t]
\centering\scriptsize
\caption{Ablations of the body and hand chain on Seamless Interaction (276 test windows, mean $\pm$ 95\% CI over 10 seeds). Each row changes one element of our final chain. ``Sampling only'': the audio is
removed at test time only. Bold: best of our rows.}
\label{tab:sl_BH_abl}
\setlength{\tabcolsep}{4pt}%
\begin{tabular}{p{4.2cm}cccccc}
\toprule
& \multicolumn{3}{c}{FGD$\downarrow$} & \multicolumn{3}{c}{closer to the
GT value is better ($\rightarrow$GT)} \\
\cmidrule(lr){2-4}\cmidrule(lr){5-7}
& body22 & upper & hands30 & \shortstack{head--speech\\r} &
\shortstack{finger\\speak/listen} & \shortstack{listening head\\shakes/min}
\\
\midrule
GT & 0.000 & 0.000 & 0.000 & 0.194 & 1.37 & 10.9 \\
\midrule
\raggedright our final chain & \textbf{0.383}{\scriptsize$\pm$.025} &
\textbf{0.207}{\scriptsize$\pm$.006} & 2.558{\scriptsize$\pm$.058} &
\textbf{0.121}{\scriptsize$\pm$.005} & \textbf{1.29}{\scriptsize$\pm$.01} &
\textbf{32.3}{\scriptsize$\pm$.4} \\
\raggedright without the arm-range and listening-head losses &
0.484{\scriptsize$\pm$.024} & 0.260{\scriptsize$\pm$.006} &
2.025{\scriptsize$\pm$.087} & 0.087{\scriptsize$\pm$.004} &
1.27{\scriptsize$\pm$.01} & 55.1{\scriptsize$\pm$.5} \\
\raggedright body module without audio (sampling only) &
0.402{\scriptsize$\pm$.020} & 0.234{\scriptsize$\pm$.005} &
\textbf{1.510}{\scriptsize$\pm$.025} & 0.005{\scriptsize$\pm$.007} &
1.12{\scriptsize$\pm$.01} & 41.1{\scriptsize$\pm$.6} \\
\raggedright hand module without audio (sampling only) &
0.383{\scriptsize$\pm$.025} & 0.207{\scriptsize$\pm$.006} &
2.566{\scriptsize$\pm$.057} & 0.121{\scriptsize$\pm$.005} &
1.16{\scriptsize$\pm$.01} & 32.3{\scriptsize$\pm$.4} \\
\bottomrule
\end{tabular}
\end{table*}

%% file: tabels/table7_seamless_face_ablation.tex
\begin{table*}[t]
\centering\small\setlength{\tabcolsep}{4pt}
\caption{Ablations of the face module on Seamless Interaction, scored on the GT body (276 test windows, mean $\pm$ 95\% CI over 10 runs). The our final face module row uses the first 10 runs of the ``Ours, GT body'' row of
Table~\mainTabSlFace{} of the main paper. Bold: best of our rows.}
\label{tab:sl_face_abl}
\resizebox{\textwidth}{!}{%
\begin{tabular}{lcccccc}
\toprule
& FRC/seed$\uparrow$ & FRD$\downarrow$ & FRVar ($\rightarrow$GT 0.985) &
S-MSE$\uparrow$ & FRDvs$\uparrow$ & TLCC$\downarrow$ (frames) \\
\midrule
GT (face) & 1.000 & 0.00 & 0.985 & -- & 4.507 & 35.5 \\
\midrule
our final face module & \textbf{0.296}{\scriptsize$\pm$.000} &
\textbf{40.83}{\scriptsize$\pm$.01} & 1.021{\scriptsize$\pm$.000} &
0.643{\scriptsize$\pm$.001} & 2.833{\scriptsize$\pm$.002} &
32.5{\scriptsize$\pm$.1} \\
without the body input & 0.153{\scriptsize$\pm$.000} &
43.23{\scriptsize$\pm$.01} & 0.925{\scriptsize$\pm$.002} &
\textbf{1.320}{\scriptsize$\pm$.003} & 2.139{\scriptsize$\pm$.004} &
\textbf{32.4}{\scriptsize$\pm$.1} \\
without audio on silent frames & 0.281{\scriptsize$\pm$.000} &
40.88{\scriptsize$\pm$.01} & \textbf{1.000}{\scriptsize$\pm$.001} &
0.629{\scriptsize$\pm$.001} & \textbf{2.855}{\scriptsize$\pm$.003} &
32.6{\scriptsize$\pm$.1} \\
\bottomrule
\end{tabular}%
}
\end{table*}